\documentclass[review,10pt,3p]{elsarticle}
\usepackage{graphicx} 
\usepackage{lineno,hyperref}
\usepackage{amsfonts,amssymb}
\usepackage{epsfig} 
\usepackage{multirow}
\usepackage{float}
\usepackage{subfig}
\usepackage{amsmath}
\usepackage{mathrsfs}
\usepackage{makecell}
\usepackage{color}
\usepackage{bm}
\usepackage{amsmath}
\usepackage{algorithm}
\usepackage{algorithmic}
\usepackage{textcomp}
\usepackage{amsmath,amssymb,amsfonts}
\usepackage{caption}
\usepackage{wrapfig}

\floatstyle{plaintop}
\restylefloat{table}

\journal{Journal of \LaTeX\ Templates}

\begin{document}	
 \begin{frontmatter}		
		\title{Joint Representation of Multiple Geometric Priors via a Shape Decomposition Model for Single Monocular 3D Pose Estimation}
        \author{Mengxi Jiang\textsuperscript{1}, Zhuliang Yu\textsuperscript{2}, Cuihua Li\textsuperscript{1}, Yunqi Lei\textsuperscript{1,*}}
       \address{\textsuperscript{1}Department of Computer Science, Xiamen  University, 361005, China\\
    	\textsuperscript{2}College of Automation Science and Engineering, South China University of Technology, Guangzhou 510640, China\\       	
	    jiangmengxi@stu.xmu.edu.cn, zlyu@scut.edu.cn, \{chli,yqlei\}@xmu.edu.cn}
		
		
        \cortext[mycorrespondingauthor]{Corresponding author: Yunqi Lei}
        \ead{jiangmengxi@stu.xmu.edu.cn, zlyu@scut.edu.cn, \{chli,yqlei\}@xmu.edu.cn}

\begin{abstract}
In this paper, we aim to recover the 3D human pose from 2D body joints of a single image. The major challenge in this task is the depth ambiguity since different 3D poses may produce similar 2D poses. Although many recent advances in this problem are found in both unsupervised and supervised learning approaches, the performances of most of these approaches are greatly affected by insufficient diversities and richness of training data. To alleviate this issue, we propose an unsupervised learning approach, which is capable of estimating various complex poses well under limited available training data. Specifically, we propose a Shape Decomposition Model (SDM) in which a 3D pose is considered as the superposition of two parts which are global structure together with some deformations. Based on SDM, we estimate these two parts explicitly by solving two sets of different distributed combination coefficients of geometric priors. In addition, to obtain geometric priors, a joint dictionary learning algorithm is proposed to extract both coarse and fine pose clues simultaneously from limited training data. Quantitative evaluations on several widely used datasets demonstrate that our approach yields better performances over other competitive approaches. Especially, on some categories with more complex deformations, significant improvements are achieved by our approach. Furthermore, qualitative experiments conducted on in-the-wild images also show the effectiveness of the proposed approach.

\end{abstract}

\begin{keyword}
	\texttt{3D Pose Estimation}\sep \texttt{Sparse Representation Model} \sep \texttt{Shape Decomposition Model}\sep \texttt{Human Body Estimation} \sep \texttt{Skeleton Recognition} \sep \texttt{Multiple Geometric Learning} 
\end{keyword}

\end{frontmatter}

\section{Introduction}
\label{sec:introduction}
In many computer vision applications, such as human-robot interaction, human action surveillance, and virtual reality (VR), etc, it is necessary to recover the 3D human pose from 2D images. However, this is a challenging task due to the lack of depth information in 2D images. To overcome this difficulty, some literature have presented typical solutions by assuming that the RGB-D image~\cite{Jung2015Random} or the multi-view image~\cite{rhodin2018learning,Shen2013Part} is available. Benefited from these supplementary data, many works have made impressive progress over the past years. However, in many real-life scenes, it is infeasible to develop specific hardware equipment for the capture of depth information or multi-view images. Thus, the ultimate goal is to recover the 3D human pose from a single RGB image which can be obtained easily. 

Lifting 2D observations of a single image to 3D space, there are two main factors leading to difficulties. Firstly, influenced by many factors, such as illuminations, cluttered backgrounds or occlusions, 2D human features are hard to locate in the image. Secondly, even ground-truth 2D body features are given, it is still a highly ill-posed problem since similar 2D projections may be projected from different 3D poses. To overcome the above difficulties, existing approaches have proposed different solutions over the past years.

The first class of approaches directly learn a mapping from image features to 3D  pose~\cite{Bo2010Twin} or search the best-matched example from an abundant of stored correspondences between 2D observations and its 3D pose~\cite{Jiang20103D, Chen20113D}. Benefited from the development of Convolutional Networks (ConvNets), researchers focus on the ConvNets-based of end-to-end learning approaches to estimate the 3D pose from the 2D image features~\cite{Li20143D, Tekin2016Structured, Katircioglu2018Learning}. Another category is  ed as the model-based method~\cite{Ramakrishna2012Reconstructing, Zhou2016Sparse}, which infers the 3D pose from 2D joints of an image by optimizing a designed objective function based on a given human body model. Recently, the 3D pose estimation problem has been addressed by a hybrid approaches~\cite{Zhou2016Sparseness, Bogo2016Keep}. In these works, 2D joints are used as human body features which are detected by existing robust 2D detectors at first~\cite{Newell2016Stacked}. Then, the model-based approaches are applied as a post-processing optimization to infer the 3D pose from detected 2D joints. Since the 2D joints are relatively easy to obtain, the model-based approaches become critical to the performance of the 3D pose estimation.

\begin{figure}[h]
	\centering
	\includegraphics[height=1.99in,width=3.4in]{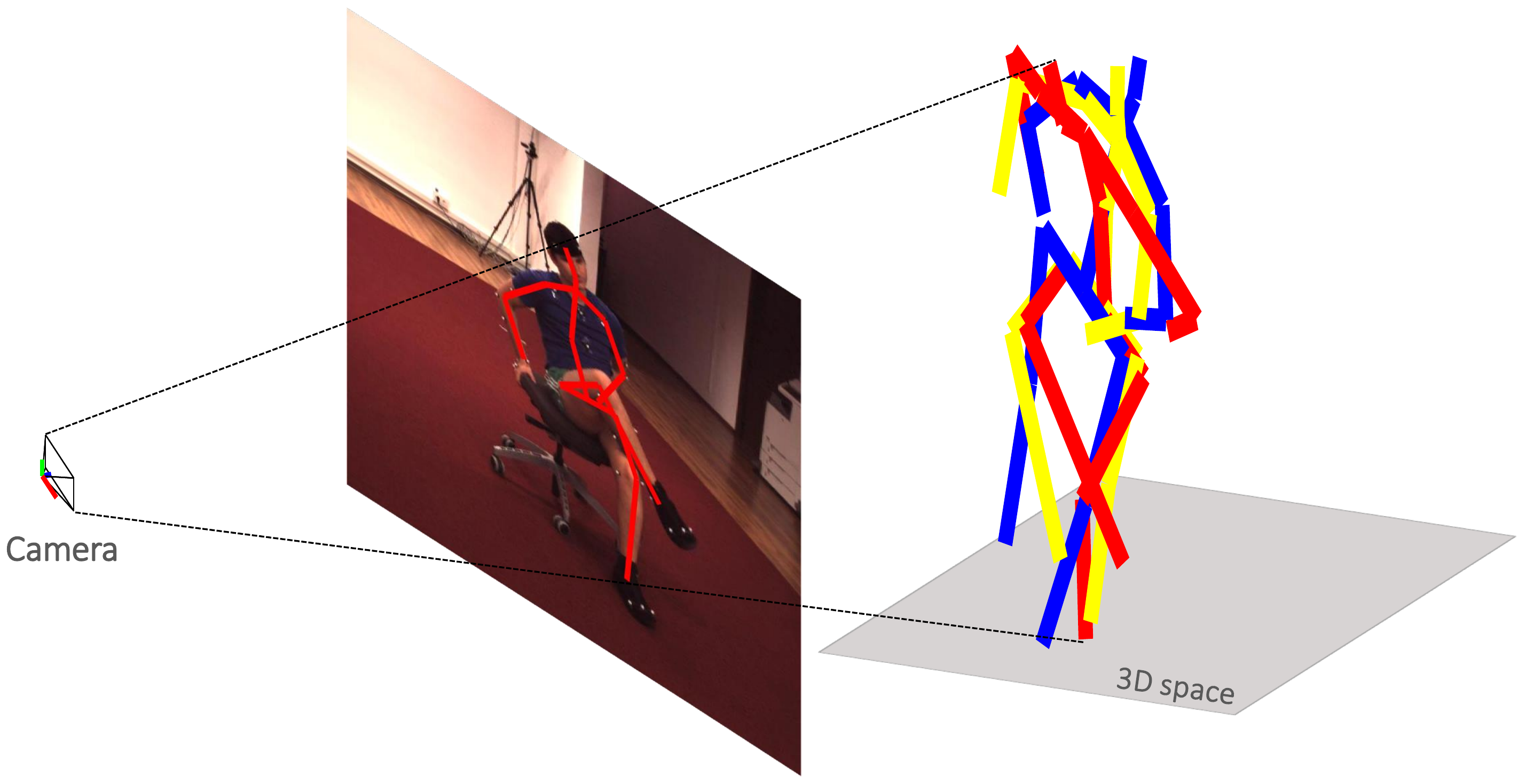}
	\caption{The example of 3D pose estimation. The pose with color red denotes the ground-truth. The 3D pose with color blue and color yellow are estimated results from standard SR model and the proposed model respectively.}
	\label{fig:lexample}
\end{figure}

Sparse representation (SR) model~\cite{Olshausen1997Sparse} is an effective model-based method to combine 2D annotations with 3D priors learned from existing Mocap dataset~\cite{CMU,Sigal2006HumanEva,Ionescu2014Human3}. Based on SR model, each 3D pose is represented as a linear combination of a few of pre-defined 3D bases. The effectiveness of the sparsity combination assumption has been validated in many SR based approaches~\cite{Zhou2016Sparse, Zhou2017MonoCap, Wang2018Robust,dou2018monocular,Xu2017Recovering,Shao2018Spatial}. However, the success of SR based algorithms highly depend on the assumption that sufficient representative 3D bases must be guaranteed for every input signal. For example, we consider an extreme case that the training data contain only ``Walk" action. In this case, the standard SR based model~\cite{Ramakrishna2012Reconstructing} fails to reconstruct the deformations of a ``Sit" pose, as the 3D pose with the color blue shown in Figure~\ref{fig:lexample}.

The main reason for this issue is that 3D bases obtained from ``Walk" training data are not representative enough to approximate the input ``Sit" pose. In addition, the sparsity assumption limits SR model to capture more implicit details from all bases. Thus, the diversities of 3D training data are needed for SR based approaches to handle various complex poses. However, rich and diverse training samples are difficult to obtain since it is almost infeasible to collect the 3D annotations data in unconstrained scenes. In addition, although several available Motion Capture (MoCap) datasets are widely used as training sources, they are collected in indoor. Such a collection environment greatly limits the diversity of subjects and actions. In order to augment the training data, some researchers propose different synthesize techniques~\cite{Chen2016Synthesizing}. However, the same distributions between these synthesized data cannot be guaranteed. As a result, the diversity requirement of training data for SR based approaches is hard to be satisfied. 

In this paper, in order to enhance the standard SR model to estimate various complex poses under available training data, we propose the following solutions. First of all, inspired by the fact that a complex deformation human pose is usually derived from a standard pose, we propose a shape decomposition model (SDM) in which a 3D pose is separated into two parts which are the global structure and the corresponding deformations. It is a natural assumption since most of the deformed poses can be considered as a standard pose added with some specific variations. The advantage of SDM is that the complex deformations of the 3D pose are explicitly modeled for the subsequent refined estimation. 

Based on SDM, for the estimation of the global structure of 3D poses, similar as the sparse assumption used in most SR based approaches, we represent the global structure as a sparse linear combination of global bases. While for deformations of 3D poses, we model pose deformations as a dense liner combination of deformation bases instead of the sparse combination. The main reason for this strategy is that different pose may have similar bends in some part of the human body, for example, the squatting and weightlifting have a similar knee bending. The dense constraint means that each 3D pose is estimated by using deformation clues from all 3D bases which may from other types of 3D training poses.

\begin{figure*}[h]
	\centering
	\includegraphics[scale=0.49]{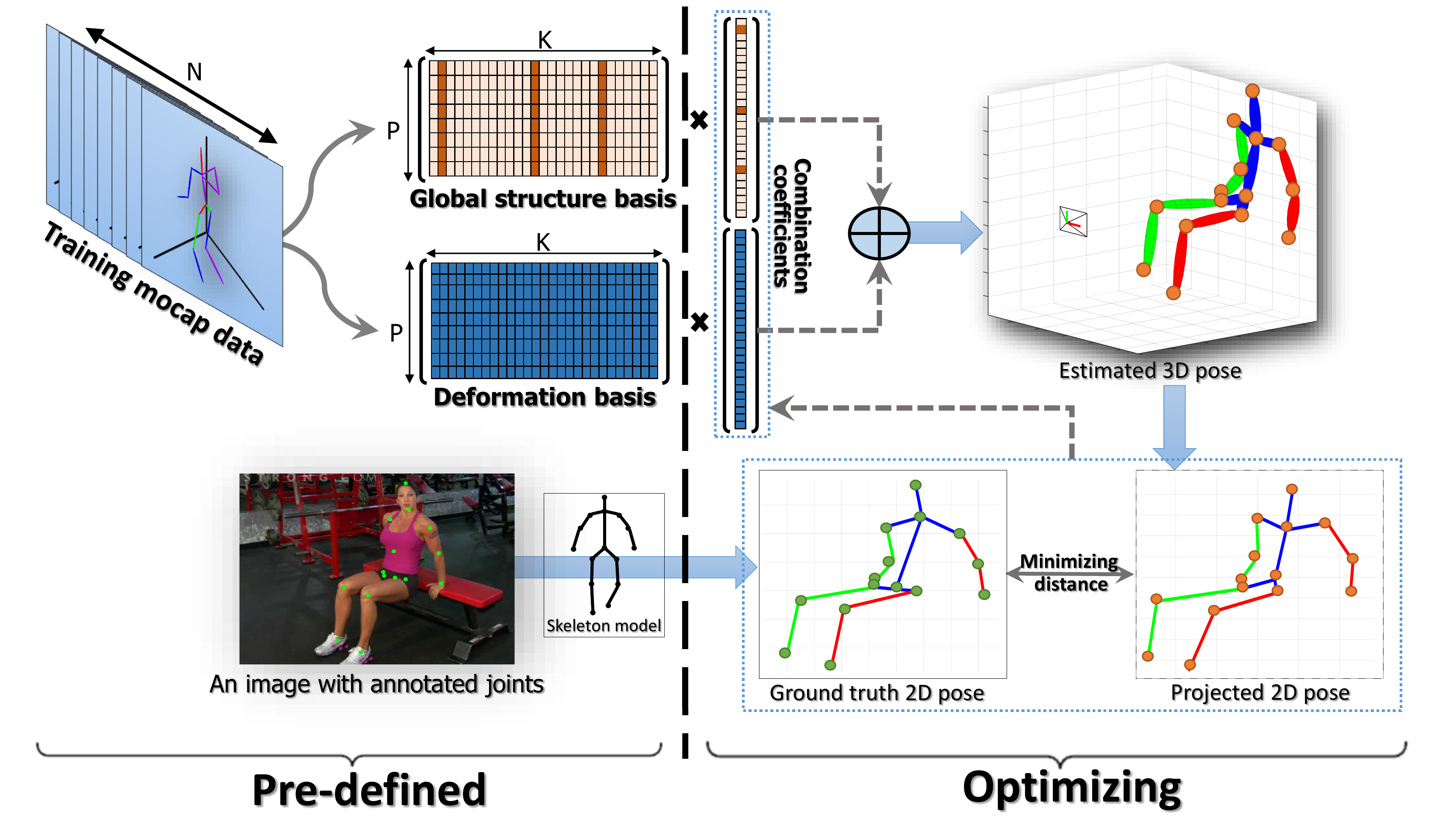}
	\caption{An overview of the proposed approach.}
	\label{fig:overview}
\end{figure*}

Further, to learn global structure and deformation dictionaries which contain a great deal of 3D bases, we propose joint learning of multiple geometric priors approach. Comparing with the single dictionary learning in most of SR based approaches, the proposed learning approach is able to obtain the dual dictionaries which captures more implicit information from the limited training data. The overview of the proposed approach is depicted in Figure~\ref{fig:overview}. Concretely, we train two sets of basis from MoCap dataset in advance. Then, a 3D pose is obtained by adding results from two sets of linear combinations of basis. Next, two sets of combined coefficients are solving by minimizing the given 2D pose of an image and the projected 2D pose (projected from the estimated 3D pose).

Our strategies are able to reconstruct a reasonable 3D pose even if the type of input pose is not included in the training data, as the 3D pose with the color yellow shown in the Figure~\ref{fig:lexample}. In general, the main contributions of this paper are summarized as follows:
\begin{itemize}
	\item Considering that a human pose can be separated as a global structure and corresponding deformations, we propose a Shape Decomposition Model (SDM) which allows us to estimate two parts of 3D pose explicitly.
	
	\item A hybrid representation strategy is presented to model two parts of the 3D pose. Specially, for global structure and deformations parts, we represent them as a sparse and dense linear combination of learned 3D bases respectively. In this collaborative representation, a complex 3D pose is estimated from coarse to fine.
		
	\item  To learn geometric priori, we propose the joint learning approach of multiple geometric priors, in which two over-complete dictionaries, including a number of 3D bases, are learned simultaneously. Our dictionary learning approach extract more implicit information from the available limited training data.
\end{itemize}

The remainder of this paper is organized as follows: the related works are reviewed in Section \ref{s1}. The problem formulation is described
in Section \ref{s2}. The proposed model and inference details are given in Section \ref{s3}. Section \ref{s4} presents the joint learning approach of dual-dictionaries. Section \ref{s5} describes experimental results and analysis. Section \ref{s6} concludes the paper.

\section{Related Work}\label{s1}

It is well known that estimating the 3D pose form a single RGB image is a very ambiguous problem. This issue can be alleviated by many solutions which can be roughly classified as the model-free and model-based methods. More recently, hybrid approaches have emerged by combining the above two classes. 

The model-free methods estimate the 3D pose from 2D observations of the image directly. Elgammal and Lee ~\cite{elgammal2004inferring} infer the 3D pose from human silhouettes by learning the representations of activity manifolds. Ankur and Bill~\cite{Agarwal20043D}apply a nonlinear regression to estimate pose from shape descriptor vectors of silhouettes. To improve the estimation performance, Sedai al.~\cite{Sedai2013Discriminative} combines shape and appearance descriptors to infer the 3D pose jointly based on a discriminative learning framework. Considering that the above approaches cannot ensure the interdependencies between outputs, Bo and Sminchisescu~\cite{Bo2010Twin} use the Kullback-Leibler divergence to match the distributions of inputs and outputs. Unlike the above learning-based approaches, Jiang al.~\cite{Jiang20103D} searched the optimal poses through millions of exemplars using the nearest neighbor scheme. As the deep convolutional networks yield significant performance in many areas, various ConvNets architectures are designed to estimate the 3D pose~\cite{Toshev2013DeepPose,Li20143D,Li2016Maximum,Morenonoguer20173D,pavlakos2017volumetric,martinez2017simple,pavlakos2018ordinal}. However, the collection of a large of paired 2D-3D correspondence for the performance of model-free techniques is still a challenge although many data augmentation approaches have been proposed~\cite{Chen2016Synthesizing,varol2017learning}.

Unlike the model-free approaches, the model-based method~\cite{Simo2012Single,Ramakrishna2012Reconstructing,Simo2013A,Zhou2016Sparse,Bogo2016Keep,Sanzari2016Bayesian,Chen20163D,Zhou2017MonoCap,Wang2018Robust} only use 3D annotations to fit their models. Based on prior knowledge, the model-based methods mainly contains two parts, the modeling and the inferring. A commonly used model is the active shape model (ASM)~\cite{Cootes1995Active}, where a 3D human pose is represented as a linear combination of 3D bases~\cite{Ramakrishna2012Reconstructing, Simo2012Single, Zhou2016Sparse, Zhou2017MonoCap, Wang2018Robust}. Ramakrishna et al.~\cite{Ramakrishna2012Reconstructing} estimate the model parameters by minimizing the projection error based on sparse representation model. Simo-Serra et al.~\cite{Simo2012Single} propagate the noise to the pose space by using a stochastic sampling strategy. Zhou et al.~\cite{Zhou2016Sparse} propose a convex formulation to solve the non-convex problems of the orthogonality constraint imposed on the objective function. Instead of using the $\ell_2$-norm to measure the reconstruction error, Want et al.~\cite{Wang2018Robust} apply the $\ell_1$-norm to minimize the projection error to improve the estimation robustness. Apart from ASM, Kostrikov et al.~\cite{Kostrikov2014Depth} apply 3D pictorial structure to estimate the 3D pose based on regression forests. Based on the mixture of pictorial structures, Radwan et al.~\cite{Radwan2014Monocular} impose kinematic and orientation constraints on the reasoning model for the self-occlusion pose estimation.

The hybrid methods combine the model-free and model-based approaches to estimate 3D human pose from an image. In the strategy, the existing generative 2D detectors~\cite{Newell2016Stacked} are only used to detect 2D features, such as 2D body joints, and then the generative approaches are applied to infer 3D pose from these features~\cite{Simo2013A, Yu2016Marker, Bogo2016Keep, Zhou2017MonoCap}. Since the literature worked on the 3D pose estimation are vast, we are unable to cover all approaches in this section, the readers interested in this topic are referred to the survey~\cite{Sarafianos20163D}.
\section{Problem Formulation}\label{s2}

\subsection{Dependence between the 2D and 3D poses} \label{s2-1}
In this paper, a human body is represented as a skeleton with $P$ joints, its 2D pose and 3D pose are described as $\bm{X}=\{\bm{j'}_i\}_{i=1}^P\in \mathbb{R}^{2 \times P}$ and 
$\bm{Y}=\{\bm{j}_i\}_{i=1}^P\in \mathbb{R}^{3 \times P}$ respectively, where $j'\in \mathbb{R}^{2}$ and $j\in \mathbb{R}^{3}$ denote joint coordinates in 2D and 3D spaces respectively. Based on the assumption of a weak perspective camera model, the projection relationship between the 2D pose and its corresponding 3D pose is modeled as 

\begin{equation}\label{1}
\bm{X}=\bm{S}\bm{R}\bm{Y}+\bm{T},
\end{equation}
where $\bm{R} \in \mathbb{R}^{3 \times 3}$ represents the rotation matrix and $\bm{T}=\bm{t}\bm{1}^\top \in \mathbb{R}^{2 \times P}$ represents the translation vector, where $\bm{t} \in \mathbb{R}^{2}$. $\bm{S}=\left[ \begin{array}{ccc}
s_x & 0 & 0\\
0 & s_y & 0\\ 
\end{array} \right]$ denotes the camera calibration matrix, where $s_x$ and $s_y$ are scaling factors in the x-axis and y-axis respectively. Given a 2D pose, the aiming of this paper is to find the 3D pose $Y$ whose projection is as equal as possible to the given 2D pose $X$. However, this is a ill-posed problem since there may be many 3D poses that satisfies Eq.~\eqref{1}. Fortunately, this problem can be alleviated by sparse representation (SR) model~\cite{Olshausen1997Sparse}.

\subsection{Sparse Representation for 3D pose estimation} \label{s2-2}

Based on SR model~\cite{Olshausen1997Sparse}, a 3D pose is represented as a sparse linear combination of a set of bases as follow: 
\begin{equation}\label{2-1}
\bm{Y}=\bm{B^*c},
\end{equation}
where $\bm{B^*}=\{\bm{B}_j\}_{j=1}^k$ is the over-completed dictionary, $\bm{B}_j\in \mathbb{R}^{3 \times P}$ represents $j$th 3D basis pose, $k$ is the size of dictionary.
$\bm{c}=[c_1,...c_k]^T$ is the sparse coefficient vector in which most of entries are expected to be zero. As a result, only a few bases in dictionary are activated to express the 3D pose.

Since the dictionary is pre-learned from the training data, the problem of estimating the 3D pose $Y$ is converted to the solving of sparse coefficient $c_i$. By combining Eq.~(\ref{1}) and Eq.~(\ref{2-1}), the standard objective function based on SR model is defined as

\begin{equation}\label{2-2}
\begin{aligned}
\min_{\bm{R},\bm{c},\bm{T}} \Vert \bm{c}\Vert_1 \quad \textbf{\emph{s.t.}} \quad \bm{X}=\bm{SR(B^*c)}+\bm{T}, 
\end{aligned}
\end{equation}
where $\Vert \cdot \Vert_1$ is $\ell_1$-norm which used to ensure the sparsity of coefficients vector $\bm{c}$. Notice that $\ell_1$-norm is the convex approximation of $\ell_0$-norm since the solving of $\ell_0$-norm is NP-hard~\cite{Candes2006Near}. 

Eq.~\eqref{2-2} is the standard SR formulation to model the task of 3D pose estimation from 2D joints of an image, which is widely used in most unsupervised learning approaches~\cite{Ramakrishna2012Reconstructing,Zhou2016Sparse,Wang2014Robust}. The sparsity constraint $\Vert \cdot \Vert_1$ in Eq.~\eqref{2-2} means that only a few bases poses are selected from the dictionary to represent a 3D pose. In addition, these activated bases are often in the same category as input 2D pose, and the more similar bases to the input signal are usually assigned higher combination weight values. This is the reason that SR model also performs well in the classification task~\cite{Shao2018Spatial}.

However, we observed that two poses with large overall differences may have similar bends in some local joints the human body, thus those unselected basis pose also are able to provide useful clues for deformation estimations of the 3D pose. Unfortunately, the standard SR model fails to capture detail deformations from unselected bases which may learned from different category training data. As a result, the estimation performance of SR based approaches is affected, especially when the type of input pose is not included in training data. To alleviate this problem, we propose our approach in the coming section.

\section{Proposed Models}\label{s3}

\subsection{Shape Decomposition Model and 3D Pose Inference} \label{s3-1}

Considering that an arbitrary 3D pose $\bm{Y}$ contains both the global structure and its corresponding detail deformations, we describe a 3D pose  as
\begin{equation}\label{3-1}
\bm{Y}=\bm{U}+\bm{V},
\end{equation}
where $\bm{U}\in \mathbb{R}^{3 \times P}$ and $\bm{V}\in \mathbb{R}^{3 \times P}$ denotes the global structure of the 3D pose and its deformations respectively. Applying the active shape model~\cite{Cootes1995Active}, we represent each part as
\begin{align}
& \bm{U}=\bm{B_u^*}\bm{c_u}, \label{eq:rel1}\\
& \bm{V}=\bm{B_v^*}\bm{c_v}, \label{eq:rel2}
\end{align}
where $\bm{B_u^*}=\{\bm{B}_{\bm{u}j}\}_{j=1}^k$and $\bm{B_v^*}=\{\bm{B}_{\bm{v}j}\}_{j=1}^k$ denote 
the global structure and deformation dictionaries respectively, whose learning strategy is described in section~\ref{s4}. $\bm{c_u} \in \mathbb{R}^{k}$ and $\bm{c_v} \in \mathbb{R}^{k}$ are corresponding combination coefficient vectors. We term Eq.~\eqref{3-1} as Shape Decomposition Model (SDM). Since the dictionaries $\bm{B_u^*}, \bm{B_v^*}$ is pre-defined, the 3D pose can be calculated by using two sets of combination coefficients $\bm{c_u^*}, \bm{c_v^*}$ which are possible to be solved by imposing suitable constraints. Thus, by combining Eq.~\eqref{1} and Eq.~\eqref{3-1}-\eqref{eq:rel2}, we formulate the optimization problem as
\begin{equation}\label{3-2}
\begin{aligned}
\min_{\bm{R},\bm{c_u},\bm{c_v}} \phi(\bm{c_u}) + \varphi(\bm{c_v}) \;\   \textbf{\emph{s.t.}} \;\ \bm{X}=(\bm{SR)(B^*_uc_u+B^*_vc_v)}+\bm{T}, 
\end{aligned}
\end{equation}
where $\phi(\cdot)$ and $\varphi(\cdot)$ are constraint terms on the coding of global structures of 3D poses and its deformations information. 

For the coding of global structure part, based on the assumption that the 3D pose lies in a small subspace~\cite{Ramakrishna2012Reconstructing, Zhou2016Sparse, Zhou2016Sparseness, Wang2018Robust}, we use $\Vert \cdot \Vert_1$ on $\bm{c_u}$ to enforce the solution sparsity. While for the coding of body deformations, considering that different types poses may have similar deformations in some local body part, we use the dense constraint $\Vert \cdot \Vert_2$ on $\bm{c_v}$ to produce a dense solution which implies that pose deformation is represented collaboratively by all bases in the dictionary $\bm{B^*_v}$. This strategy is successfully applied in the face recognition task \cite{Xudong2015Sparse}. In addition, considering that Gaussian noises may exist in Eq.~\eqref{3-1}, we relax the equality constraint of Eq.~\eqref{3-2}. As a result, Eq.~\eqref{3-2} is rewritten as following Lagrangian form

\begin{equation}\label{3-3}
\begin{aligned}
\min_{\bm{R},\bm{c_u},\bm{c_v}} \frac{1}{2}\Vert \bm{X}-(\bm{SR)(B^*_uc_u+B^*_vc_v)}+\bm{T} \Vert_2^2
+\alpha\Vert \bm{c_u} \Vert_1 + \beta\Vert \bm{c_v} \Vert_2^2,
\end{aligned}
\end{equation}
where $\alpha$ and $\beta$ are balance factors for each minimization term. 

For further simplicity, we remove the translation vector $\bm{T}$ in  ~\eqref{3-3} by centralizing the data, and we introduce a new variable $\bm{R^*} \in \mathbb{R}^{2 \times 3}$ to denote the product of $\bm{S}$ and $\bm{R}$. Ultimately, the final objective function is given as 
\begin{equation}\label{3-4}
\begin{aligned}
\min_{\bm{R^*},\bm{c_u},\bm{c_v}} \frac{1}{2}\Vert \bm{X}-\bm{R^*(B^*_uc_u+B^*_vc_v)} \Vert_2^2 
+\alpha\Vert \bm{c_u} \Vert_1 + \beta\Vert \bm{c_v} \Vert_2^2,
\end{aligned}
\end{equation}

\subsection{Optimization}\label{s3-5}

To solve the optimization problem of the model \eqref{3-4}, we apply the Alternating Direction Method of Multipliers (ADMM)~\cite{boyd2011distributed}, in which the variables $\bm{R^*},\bm{c_u}, \bm{c_v}$ are updated alternatively. 

\textbf{Updating $\bm{R^*}$:} We fix $\bm{c_u}, \bm{c_v}$, $\bm{R^*}$ is obtain by minimizing the following subproblem: 
\begin{equation}\label{5-1}
\begin{aligned}
\bm{R^*}=\min_{\bm{R^*}} \frac{1}{2}\Vert \bm{X}-\bm{R^*}(\bm{B_u^*}\bm{c_u}+\bm{B^*_v}\bm{c_v}) \Vert_2^2.
\end{aligned}
\end{equation}

The problem~\eqref{5-1} is a differentiable function which can be solved by the manifold optimization toolbox~\cite{Boumal2013Manopt}.

\textbf{Updating $\bm{c_u}$:} By fixing $\bm{R^*}$ and $\bm{c_v}$, the subproblem about $\bm{c_u}$ is given as
\begin{align}\label{5-2}
\bm{c_u}&=\min_{\bm{c_u}} \frac{1}{2}\Vert \bm{X}-\bm{R^*}(\bm{B_u^*}\bm{c_u}+\bm{B^*_v}\bm{c_v})\Vert_2^2 +\alpha\Vert \bm{c_u} \Vert_1.
\end{align}

The problem~\eqref{5-2} is convex, thus it can be solved by $\ell_1$-norm minimization solvers, such as Accelerated Proximal Gradient (APG) algorithm~\cite{Nesterov2013Gradient}.

\textbf{Updating $\bm{c_v}$:} Fixing $\bm{R^*}$ and $\bm{c_u}$, the $\bm{c_v}$ is solved by following optimization problem:
\begin{align}\label{5-3}
\bm{c}_v &= \min_{\bm{c_v}} \frac{1}{2}\Vert \bm{X}-\bm{R^*}(\bm{B_u^*}\bm{c}_u+\bm{B_v^*}\bm{c_v}) \Vert_2^2 + \beta\Vert\bm{c_v}\Vert_2^2.
\end{align}

The sub-problem~\eqref{5-3} consists of two differentiable functions, it exists a closed-form solution since it is a convex issue. By introducing an auxiliary variable $\bm{Z}=\{\bm{R^*B}_{\bm{v}i}\}_{i=1}^k$, we present the solution of $\bm{c_v}$ as
\begin{equation}\label{5-4}
\bm{c_v}=(\bm{Z}^T\bm{Z}+\beta\textbf{I})^{-1}\bm{Z}^T(\bm{X}-\bm{R^*}(\bm{B_u^*}\bm{c_u})).
\end{equation}
where $\textbf{I}$ is identity matrix. The completed reconstruction approach based on SDM is summarized in Algorithm 1.
\begin{algorithm}[h]\small
	\caption{ADMM to solve the problem Eq.\ref{3-4}.}
	\label{alg:algorithm}
	\textbf{Input}: $\bm{X},\bm{B_u^*,B_v^*}$ \quad \textcolor[RGB]{88,87,86}{//The input 2D joints, the main pose and deformations dictionaries.} \\
	\textbf{Parameter}: $\tau, \alpha, \beta$ \quad \textcolor[RGB]{88,87,86}{//The tolerance for the object value, the balance factors.} \\
	\textbf{Output}: $\bm{Y}$
	\begin{algorithmic}[1] 
		\STATE initialize $\bm{c_u}, \bm{c_v},\bm{R^*}, \tau, \alpha, \beta$.
		\WHILE{$\Vert\bm{r}\Vert_2>\tau$ or $\ell<\ell_{max}$}
		\STATE \emph{\textbf{update}} $\bm{R^*}$ \emph{by Eq.} (\ref{5-1}).
		\STATE \emph{\textbf{update}} $\bm{c_u}$ \emph{by Eq.} (\ref{5-2}).
		\STATE \emph{\textbf{update}} $\bm{c_v}$ \emph{by Eq.} (\ref{5-4}).
		\STATE \emph{\textbf{calculate}} $\bm{r}=\bm{X}-\bm{R^*}(\bm{B_u^*}\bm{c_u}+\bm{B_v^*}\bm{c_v})$. \textcolor[RGB]{88,87,86}{// the reconstruction residual.}
		\STATE \emph{\textbf{update}} $\ell=\ell+1$. \quad \textcolor[RGB]{88,87,86} {// iteration count.} 
		\ENDWHILE
		\STATE \emph{\textbf{calculate}} $\bm{Y}=\bm{B_u^*}\bm{c_u}+\bm{B_v^*}\bm{c_v}$; 
	\end{algorithmic}
\end{algorithm}

\subsection{Dictionary Learning}\label{s4}

The global structure dictionary $\bm{B}^*_u$ and the deformations dictionary $\bm{B}^*_v$ in Eq.~\eqref{3-4} should be properly learned to provide powerful priors for the 3D pose estimation. Given a set of training set $\bm{D^*}=[\bm{D}_1,...\bm{D}_m,...\bm{D}_N]$, we apply SDM on each training pose $\bm{D}_m$ which is formulated as
\begin{equation}
\begin{aligned}\label{4-3}
\bm{D}_m&=\bm{U}_m+\bm{V}_m,\\
& =\bm{B_u^*}\bm{c}_{\bm{u}m}+\bm{B_v^*}\bm{c}_{\bm{v}m},
\end{aligned}
\end{equation}
where $\bm{D}_m$ is the $m$-th 3D pose in training set. $\bm{U}_m$, $\bm{V}_m$ are the global structure and deformations of the $m$-th 3D training pose. $N$ is the size of training set. Similar to Eq.~\eqref{3-4}, by introducing the sparsity and density constraints, the object function of dictionary learning is presented as
\begin{equation}\label{4-4}\small
\begin{aligned} 
\min_{\{\bm{B_u},\bm{B_v}\}_1^k,\bm{C_u},\bm{C_v}} \sum_{m=1}^N  \frac{1}{2} \Vert \bm{D}_m-\sum_{j=1}^k (\bm{B}_{\bm{u}j}c_{ujm}+\bm{B}_{\bm{v}j}c_{vjm})\Vert_2^2 
+ \gamma\Vert \bm{C_u} \Vert_1 + \eta\Vert \bm{C_v} \Vert_2^2,
\end{aligned}
\end{equation}
where $c_{ujm}$ and $c_{vjm}$ denote the $j$-th sparsity and dense coefficient vectors respectively for the $m$-th training pose. $\bm{C_u}=\{\bm{c}_{\bm{u}m}\}_{m=1}^N$ and  $\bm{C_v}=\{\bm{c}_{\bm{v}m}\}_{m=1}^N$ are the combination matrices by stacking coefficient vectors $\bm{c_u}=[c_{u1},...,c_{uk}]^\top$ and $\bm{c_v}=[c_{v1},...,c_{vk}]^\top$ of all training poses respectively. $\gamma$ and $\eta$ are balance parameters for sparse and dense representation terms respectively. The formulation \eqref{4-3} can also be solved by ADMM scheme. Given training data $\bm{D^*}$, we alternately update the $\bm{B^*_u},\bm{C_u},\bm{B^*_v},\bm{C_v}$ while fixing the others. The algorithm is presented in Appendix A.

\section{Experiments}\label{s5}

In this section, we evaluate the proposed approach on the task of estimating the 3D pose from 2D joints of an image. In the following subsections, the experimental settings are presented in Section~\ref{s5-1}. The parameter sensitivity experiments are reported in Sections~\ref{s5-6}. Quantitative results and analysis conducted on different datasets are provided in Section~\ref{s5-2}, \ref{s5-3}, and \ref{s5-4}. Qualitative results are given in~\ref{s5-5}.
\subsection{Experimental settings}\label{s5-1}
\subsubsection{Evaluation Dataset and Protocol} 
The proposed approach is evaluated on four publicly available datasets, they are Human3.6M~\cite{Ionescu2014Human3}, HumanEva-I~\cite{Sigal2006HumanEva}, CMU Mocap~\cite{CMU}, and MPII Human Pose~\cite{Andriluka20142D}. Specially, the last one is used for the qualitative evaluation, and others for the quantitative evaluations. Notice that we only use the 3D pose date for training rather than paired 2D-3D data with images in most supervised approaches.~\cite{Tekin2016Direct, Zhou2016Deep,martinez2017simple}

\textbf{Human3.6M}: This is a large scale dataset that contains synchronized videos with corresponding 2D-3D human poses of \textit{11} subjects. Each subject performs \textit{15} scenarios, including sitting, smoking, and eating, etc. These processes are captured by an indoor MoCap system with \textit{15} sensors (video and motion cameras, time-of-flight sensor). The evaluation data were downsampled from \textit{50}fps to \textit{10}fps. The data from subjects (S1, S5, S6, S7, S8) was used for training, and (S9, S11) for testing. This is the standard evaluation protocol which is applied in most approaches~\cite{Zhou2016Sparseness, Tekin2016Structured, Li2016Maximum, Zhou2017MonoCap}.

\textbf{HumanEva-I}: Compared to Human3.6M dataset, HumanEva-I dataset provides the smaller synchronized sequences and corresponding 2D-3D annotations. Specifically, six actions, such as jogging, catching, boxing, etc., are performed under seven camera views by four subjects. Following the same evaluation protocols in most  literature~\cite{Radwan2013Monocular, Wang2014Robust, Simo2013A, Kostrikov2014Depth, Bo2010Twin, Yasin2016A, Zhou2017MonoCap, Katircioglu2018Learning}, we evaluated our approach on categories of "Walking" and "Jogging" performed by subjects S1, S2, and S3 from "validation" set.

\textbf{CMU Mocap}: This dataset contains \textit{30} different actions performed by \textit{144} subjects, and collects more than three million 3D human poses. For the evaluation, eight motions (basketball, dancing, boxing, jumping, running, climbing and walking) are collected from different subjects. In our evaluation, six sequences are collected from different motions respectively, where three sequences are used as training sets and the remaining three sequences are treated as test sets. To generate input 2D pose, we simulate a 360-degree rotating orthogonal camera to projected the 3D pose into the 2D space.

\textbf{MPII}: This dataset is used for the qualitative evaluation. In 2D pose estimation works, MPII is a widely used dataset which is consists of about \textit{25}k images and \textit{40}k 2D poses. These 2D poses are obtained from different camera viewpoints and exist occluded or truncated case. We select different motions and viewpoints to evaluate the effectiveness of the proposed approach.

\subsubsection{Evaluation Metrics}
By assuming that the 2D joints are given, the proposed approach recover the 3D pose from given 2D joints. Two evaluation metrics are considered in experiments. The first is \textbf{per joint 3D error} which measures the mean euclidean distance between the joints of the estimated 3D pose and the ground truth 3D pose. This metric is formulated as follow
\begin{equation}\label{12}
\bm{err_p}(\bm{\hat{Y}},\bm{Y}_{g}) = \frac{1}{N}\sum_{i=1}^N (\Vert \bm{\hat{y}}_i-\bm{y}_i \Vert_2 ),
\end{equation}
where $\bm{\hat{Y}}$ and $\bm{Y}_{g}$ represent the estimated 3D pose and the ground-truth 3D pose respectively.  The second is \textbf{3D estimation error} which also calculates the euclidean distance between two 3D poses but after a rigid alignment.
\subsubsection{Implementation Details}

The Algorithm \ref{alg:diclearn} was used to learn the global structure and deformation dictionaries. We set both dictionaries size to $K= \{258, 168, 158\}$ for Human3.6M, HumanEva-I, and CMU Mocap respectively. The hyper-parameters $\gamma, \eta$ in Eq.~\eqref{4-4} were fixed in $\{0.01, 1\}$. For all evaluated dataset, notice that we use all training examples to train our dictionaries that does not distinguish specific actions.

For the 3D pose estimation, we set hyper-parameters $\alpha, \beta$ in the proposed model Eq.~\eqref{3-4} to $\{\alpha=0.4, \beta=20\}$, $\{\alpha=0.1, \beta=10\}$, and $\{\alpha=1.5, \beta=20\}$ for Human3.6M, HumanEva-I, and CMU Mocap respectively. For these hyper-parameters, we perform ablation analyses in the coming subsection (\ref{s5-6}). 

For model variables (rotation matrix $\bm{R}$ and combination coefficient vectors $\bm{c_u}, \bm{c_v}$) in Eq.~\eqref{3-4}, we initialize them as an identity matrix, zero vectors. To evaluate the influence of initialization, we also use a convex relaxation algorithm~\cite{Zhou20153D} to initialize rotation matrix $\bm{R}$. In evaluation comparision, the proposed approach with a fine initialization is named as "Our+refined". In addition, in all experiments, the maximum iterations $L_{max}$ and the convergence tolerance $\tau$ were set to \textit{10,000} and $10^{-6}$ respectively.

\subsection{Parameter sensitivity}\label{s5-6}

In the proposed model \eqref{3-4}, hyper-parameters $\alpha$ and $\beta$ are used to balance the trade-off of each optimization term. To study the performance of our approach as these hyper-parameters change, we conduct experiments of our approach using different hyper-parameters $\alpha$ and $\beta$ on three datasets. For each dataset, we selected \textit{100} examples randomly for the evaluation. 

\begin{figure}[h]
	\centering	
	\subfloat[]{\label{Fig:diffsc}
		\includegraphics[scale=0.38]{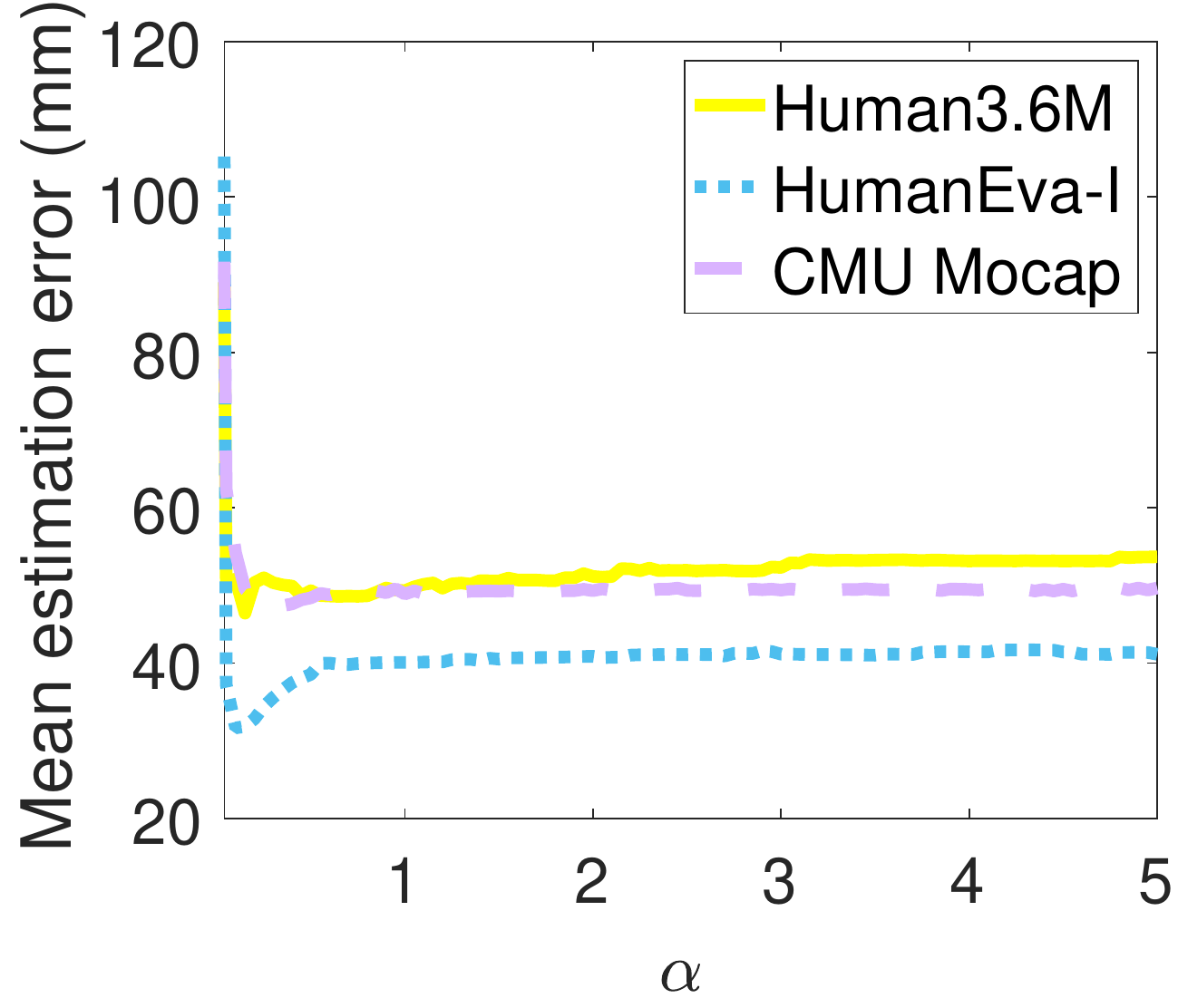}} 
	\subfloat[]{\label{Fig:diffdc}
		\includegraphics[scale=0.38]{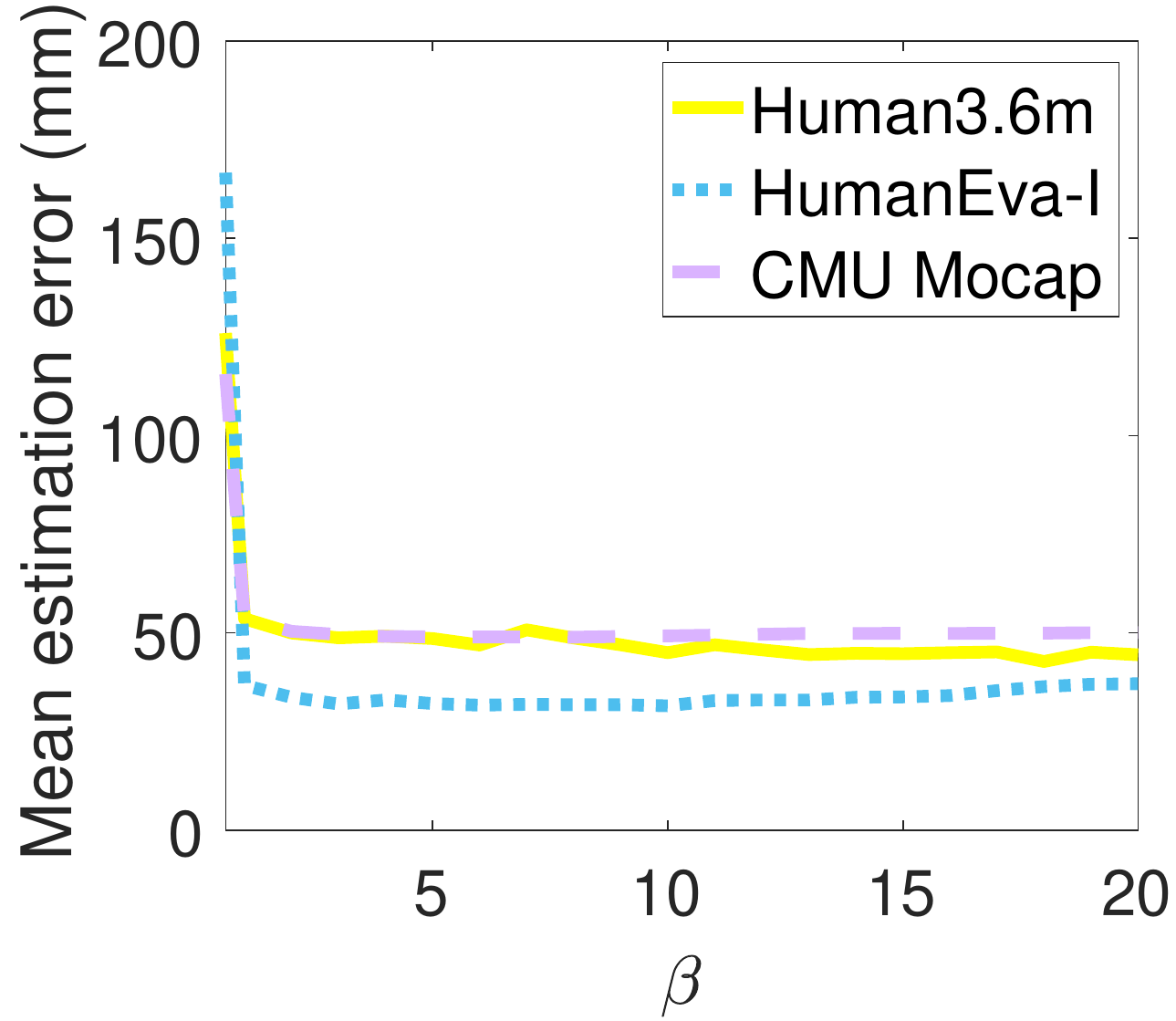}}
	\caption{The sensitivity to model hyper-parameters on different datasets. The mean estimation errors versus model hyper-parameter $\alpha$ and $\beta$ are shown, respectively.}
	\label{Fig:para_sensitty}		
\end{figure}

First of all, we perform hyper-parameter ablation analysis on parameter $\alpha$ that controls the sparsity of sparse representation. In experiments, the value of $\alpha$ is varied in the range [\textit{0, 5}] while $\beta$ is fixed to \textit{18}, \textit{10} and \textit{8} on Human3.6m, HumanEva-I, and CMU Mocap datasets respectively. As shown in Figure \ref{Fig:para_sensitty}(a),  the estimation error decreases rapidly when $\alpha$ becomes nonzero. This indicates that the sparsity is important for reconstruction performance. As $\alpha$ increases, the performance becomes very stable across all datasets. However, we find that the curve of HumanEva-I has a valley when $\alpha$ is equal to $0.1$.

Further, the ablation analysis is also conducted on $\beta$ which handles the intensity of dense representation. The value of $\beta$ is varied in the range [\textit{0, 20}] while $\alpha$ is fixed to \textit{0.15}, \textit{0.1} and \textit{1}. Similar observation is made in the ablation experiment on $\beta$, as shown in Figure \ref{Fig:para_sensitty}(b). As $\beta$ increases from zeros, the estimation errors begin to fall sharply and tend to stabilize. This indicates the importance of the dense constraint for the proposed model.  In general, the solution of our model is not very sensitive to $\alpha$ and $\beta$ when the values are in the proper range.

\subsection{Quantitative Evaluation on Human3.6M Dataset}\label{s5-2}

The first comparative study is conducted on Human3.6M dataset. In this dataset, the per joint 3D error is widely used as the performance metric in most works, while the results of estimation error are also reported in some literature. Thus, for a fair comparison, the performance from both two metrics are evaluated respectively in this paper. To evaluate the effectiveness of our approach, we compare representative works of both unsupervised and supervised learning approaches in the literature, which are organized into two groups. The result from our approach after a refined initialization is also presented, named as "\textbf{Ours+refined}". All compared figures are obtained from original papers except for the works with the marker $\dagger$ that means the results are obtained from \cite{Bogo2016Keep}.

First of all, the performances of the proposed approaches and representative works are reported in Table~\ref{table_human36m1} in terms of per joint 3D error. As seen in the table,  compared to unsupervised learning works, our approach (\textbf{Ours}) achieves the best performance in most cases and more than \textit{12}\% improvement on average. The mean accuracy is further boosted more than \textit{23}\% when we use the better initialization. Even comparison to supervised learning works, our approaches are still comparable. Specially, our approach (\textbf{Ours+refined}) outperforms most supervised learning algorithms~\cite{Tekin2016Direct,Zhou2016Deep,tekin2017learning,Nie2017Monocular,tome2017lifting,rogez2017lcr} on average. 

\begin{table}[H]\scriptsize
	\centering
	\renewcommand\arraystretch{0.7}
	\setlength{\tabcolsep}{0.3 mm}
	\begin{tabular}{lcccccccccccccccc}	
		\hline
		Methods   & Direct. & Discuss & Eat & Greet & Phone & Photo & Pose & Buy & Sit & SitDown & Smoke & Wait & WalkDog & Walk & WalkPair & Avg \\    		\hline    
		\multicolumn{17}{c}{\textit{\textbf{The supervised learning approaches}}}   \\
		\hline			                             
		Tekin et al.~\cite{Tekin2016Direct}    & \textit{102.4}   & \textit{147.2}    & \textit{88.8}   & \textit{125.3} & \textit{118.0} & \textit{182.7} & \textit{112.4} & \textit{129.2} & \textit{138.9}   & \textit{224.9}    & \textit{118.4}  & \textit{138.8} & \textit{126.3} & \textit{55.1}  & \textit{65.8}  & \textit{125.0} \\
		Zhou et al.~\cite{Zhou2016Deep}      & \textit{91.8}    & \textit{102.4}    & \textit{96.7}   & \textit{98.8}  & \textit{113.4} & \textit{125.2} & \textit{90.0}  & \textit{93.8}  & \textit{132.2}   & \textit{159.0}    & \textit{107.0}  & \textit{94.4}  & \textit{126.0} & \textit{79.0}  & \textit{99.0}  & \textit{107.3} \\
		Tekin et al.~\cite{tekin2017learning}    & \textit{85.0}    & \textit{108.8}     & \textit{84.4}   & \textit{98.9}  & \textit{119.4}  & \textit{95.7}  &\textit{98.5}  & \textit{93.8}   & \textit{73.8}    & \textit{170.4}    &\textit{85.1}  & \textit{117.0}  & \textit{113.7}  & \textit{62.1}  & \textit{94.8}  & \textit{100.1}   \\
		Nie et al.~\cite{Nie2017Monocular}  & \textit{90.1}    & \textit{88.2}     & \textit{85.7}   & 95.6  & \textit{103.9} & \textit{103.0} & \textit{92.4}  & \textit{90.4} & \textit{117.9}   & \textit{136.4}    & \textit{98.5}   & \textit{94.4}  & \textit{90.6}  & \textit{86.0}  & \textit{89.5}  & \textit{97.5}    \\
		Tome et al.~\cite{tome2017lifting}  & \textbf{\textit{65.0}}    & \textit{73.5}     & \textit{76.8}   & \textit{86.4}  & \textit{86.3}  & \textit{110.7} & \textit{68.9}  & \textit{74.8}   & \textit{110.2}   & \textit{173.9}    & \textit{85.0}   & \textit{85.8}  & \textit{86.3}  & \textit{71.4}  & \textit{73.1}  & \textit{88.4} \\
		Rogez et al.~\cite{rogez2017lcr}    & \textit{76.2}    & \textit{80.2}    & \textit{75.8}   & \textit{83.3}  & \textit{92.2}  & \textit{105.7} & \textit{79.0}  & \textit{71.7}  & \textit{105.9}   & \textit{127.1}    & \textit{88.0}  & \textit{83.7}  & \textit{86.6}  & \textit{64.9}  & \textit{84.0}  & \textit{87.7} \\
		Pavlakos et al.~\cite{pavlakos2017volumetric}   & \textit{67.4}    & \textbf{\textit{71.9}}    & \textbf{\textit{66.7}}   & \textit{\textbf{69.1}}  & \textbf{\textit{72.0}}  & \textit{77.0}  & \textbf{\textit{65.0}}  & \textit{68.3}   & \textit{83.7}    & \textit{\textbf{96.5}}     & \textbf{\textit{71.7}}   & \textit{\textbf{65.8}}  & \textbf{\textit{74.9}}  & \textit{\textbf{59.1}}  & \textbf{\textit{63.2}}  & \textit{71.9}\\
		Zhou et al.~\cite{zhou2017towards}   & \textit{\textbf{54.8}}    & \textit{\textbf{60.7}}     & \textit{\textbf{58.2}}   & \textit{71.4}  & \textit{\textbf{62.0}}  & \textit{\textbf{65.5}}  & \textit{\textbf{53.8}}  & \textit{\textbf{55.6}} & \textit{\textbf{75.2}}    & \textit{111.6}    & \textit{\textbf{64.2}}   & \textit{66.1}  & \textit{\textbf{51.4}}  & \textit{63.2}  & \textit{\textbf{55.3}}  & \textit{\textbf{64.9}}\\
		\hline	
		\multicolumn{17}{c}{\textit{\textbf{The unsupervised learning approach}}}   \\
		\hline	
		Ionescu et al.~\cite{Ionescu2014Human3}   & \textit{132.7}   & \textit{183.6}   & \textit{132.4}   & \textit{164.4} & \textit{162.1} & 205.9 & \textit{150.6} & \textit{171.3}  & \textit{151.6}   & \textit{243.0}    & \textit{162.1} & \textit{170.7} & \textit{177.1} & \textit{96.6}  & \textit{127.9}  & \textit{162.1} \\
		Yu et al.~\cite{Yu2016Marker}  & \textit{\textbf{85.1}}    & \textit{112.7}    & \textit{104.9}  & \textit{122.1} & \textit{139.1} & \textit{135.9} & \textit{105.9} & \textit{166.2}   & \textit{117.5}   & \textit{226.9}    & \textit{120.0}  & \textit{117.7} & \textit{137.4} & \textit{99.3}  & \textit{106.5} & \textit{126.5}\\
		Chen et al.~\cite{chen20173d}   & \textit{89.9}    & \textit{97.6}     & \textit{90.0}   & \textit{107.9} & \textit{107.3} & 139.2 & \textit{93.6}  & \textit{136.1}  & \textit{133.1}   & \textit{240.1}    & \textit{106.7}  & \textit{106.2} & \textit{114.1} & \textit{\textbf{87.0}}  & \textit{\textbf{90.6}}  & \textit{114.2} \\
		\textbf{Ours}  &  \textit{103.6} & \textit{101.7} & \textit{83.6} & \textit{106.3} &  \textit{86.1} &  \textit{109.3} &   \textit{96.1} &   \textit{114.5} &  \textit{97.0} &   \textit{138.3} & \textit{83.7} &    \textit{93.7} & \textit{98.4} &  \textit{100.0} &   \textit{112.0} &  \textit{99.5} \\ 
		\textbf{Ours+refined}  &  \textit{93.1} &     \textit{\textbf{85.4}} &    \textit{\textbf{72.4}} &   \textit{\textbf{95.6}} &    \textit{\textbf{80.3}} &  \textit{\textbf{100.2}} &  \textit{\textbf{90.5}} &   \textit{\textbf{90.3}} &        \textit{\textbf{74.0}} & \textit{\textbf{114.0}} &   \textit{\textbf{76.0}} &    \textit{\textbf{87.5}} & \textit{\textbf{86.7}} &   \textit{87.8} &   \textit{97.6} &   \textit{\textbf{87.1}} \\
		\hline  
	\end{tabular}
	\caption{The result of per joint 3D error (mm) on Human3.6m dataset.}
	\label{table_human36m1}
\end{table}

Moreover, experimental results of mean estimation error from our approaches and compared works are shown in Table~\ref{table_human36m2}. Similar to the observation on the per joint 3D error metric, there is a wide performance margin between the proposed approach and compared unsupervised learning approaches. Especially, our approach with refined initialization (\textbf{Ours+refined}) consistently outperforms all considered unsupervised and supervised learning approaches by more than \textit{31}\% and \textit{8}\% improvements on average respectively. In addition, the proposed approach without refined initialization (\textbf{Ours}) still achieves the best performance than all unsupervised and many supervised learning approaches \cite{Lin2017Recurrent,Morenonoguer20173D,pavlakos2017volumetric} by more than \textit{25}\% and \textit{15}\%  on average respectively.

\begin{table}[H]\scriptsize
	\centering
	\renewcommand\arraystretch{0.7}	
	\setlength{\tabcolsep}{0.13 mm}
	\begin{tabular}{lcccccccccccccccc}
		\hline
		Methods    & Direct. & Discuss & Eat & Greet & Phone & Photo & Pose & Buy & Sit & SitDown & Smoke & Wait & WalkDog & Walk & WalkPair & Avg \\
		\hline
		\multicolumn{17}{c}{\textit{\textbf{The supervised learning approach}}}     \\
		\hline 
		Lin et al.~\cite{Lin2017Recurrent}    &\textit{58.0}  & \textit{68.2}   & \textit{63.3} & \textit{65.8}  & \textit{75.3}  & \textit{93.1} & \textit{61.2} &  \textit{65.7} &\textit{98.7}    &\textit{127.7}     & \textit{70.4}  & \textit{68.2}  & \textit{72.9} & \textit{50.6}  & \textit{57.7}  & \textit{73.1}  \\
		Morenonoguer et al.~\cite{Morenonoguer20173D}    & \textit{66.1}    & \textit{61.7}     & \textit{84.5}   & \textit{73.7}  & \textit{65.2}  & \textit{67.2}  & \textit{60.9}  & \textit{67.3}  & \textit{103.5}   & \textbf{\textit{74.6}}     & \textit{92.6}   & \textit{69.6}  & \textit{71.5}  & \textit{78.0}  & \textit{73.2}  & \textit{74.0}  \\				
		Pavlakos et al.~\cite{pavlakos2017volumetric}    & \textit{47.5}    & \textit{50.5}     & \textit{48.3}   & \textit{49.3}  & \textbf{50.7}  & \textit{\textbf{55.2}}  & \textit{46.1}  & \textit{48.0}   & \textit{61.1}    & \textit{78.1}  & \textit{51.1}   & \textit{48.3}  & \textit{52.9}  & \textit{41.5}  & \textit{46.4}  & \textit{51.9}\\
		Martinez et al.~\cite{martinez2017simple}    & \textit{\textbf{39.5}}    & \textit{\textbf{43.2}}  & \textit{\textbf{46.4}}   & \textit{\textbf{47.0}}  & \textit{51.0}  & \textit{56.0}  & \textit{\textbf{41.4}}  & \textit{\textbf{40.6}}  & \textit{\textbf{56.5}}    & \textit{\textbf{69.4}} & \textit{\textbf{49.2}}  & \textit{\textbf{45.0}}  & \textit{\textbf{49.5}}  & \textit{\textbf{38.0}}  & \textit{\textbf{43.1}}  & \textit{\textbf{47.7}}  \\
		\hline
		\multicolumn{17}{c}{\textit{\textbf{The unsupervised learning approach}}}     \\
		\hline	
		Akhter et al.~\cite{Akhter2015Pose} $\dagger$ & \textit{199.2}   & \textit{177.6}    & \textit{161.8}  & \textit{197.8} & \textit{176.2} & \textit{186.5} & \textit{195.4} & \textit{167.3} & \textit{160.7}   & \textit{173.7}    & \textit{177.8}  & \textit{181.9} & \textit{176.2} & \textit{198.6} & \textit{192.7} & \textit{181.1} \\
		Ramakrishna et al.~\cite{Ramakrishna2012Reconstructing} $\dagger$& \textit{137.4}   & \textit{149.3}    & \textit{141.6}  & \textit{154.3} & \textit{157.7} & \textit{158.9} & \textit{141.8} & \textit{158.1} & \textit{168.6}   & \textit{175.6}    & \textit{160.4}  & \textit{161.7} & \textit{150.0} & \textit{174.8} & \textit{150.2} & \textit{157.3}\\
		Zhou et al.~\cite{Zhou2016Sparseness} & \textit{87.4}   & \textit{109.3}    & \textit{87.1}   & \textit{103.2} & \textit{116.2} & \textit{143.3} & \textit{106.9} & \textit{99.8}  & \textit{124.5}   & \textit{199.2}    & \textit{107.4}  & \textit{118.1} & \textit{114.2} & \textit{79.4}  & \textit{97.7} & \textit{113.0} \\
		Zhou et al.~\cite{Zhou2016Sparse} $\dagger$ \cite{Zhou2016Sparse}     & \textit{99.7}    & \textit{95.8}     & \textit{87.9}   & \textit{116.8} & \textit{108.3} & \textit{107.3} & \textit{93.5}  & \textit{95.3}   & \textit{109.1}   & \textit{137.5}    & \textit{106.0}  & \textit{102.2} & \textit{106.5} & \textit{110.4} & \textit{115.2} & \textit{106.7} \\
		Sanzari et al.~\cite{Sanzari2016Bayesian}    & \textit{48.8}  & \textit{56.3} & \textit{96.0} & \textit{84.8} & \textit{96.5} &\textit{105.6} & \textit{66.3}  & \textit{107.4} &\textit{116.9} & \textit{129.6} &\textit{97.8}   & \textit{65.9} & \textit{130.5} & \textit{92.6} & \textit{102.2}  & \textit{93.2}  \\
		Bogo eta l.~\cite{Bogo2016Keep}   & \textit{62.0}    & \textit{60.2}     & \textit{67.8}   & \textit{76.5}  & \textit{92.1}  & \textit{77.0}  & \textit{73.0}  & \textit{75.3}  & \textit{100.3}   & \textit{137.3}    & \textit{83.4}   & \textit{77.3}  & \textit{86.8}  & \textit{79.7}  & \textit{87.7}  & \textit{82.3} \\
		Wang et al.~\cite{wang20193d}   & \textit{\textbf{50.0}}    & \textit{60.0} & \textit{54.7} & \textit{56.6}  & \textit{65.7}  & \textit{78.6}  & \textit{52.7}  & \textit{54.8}  & \textit{85.9}   & \textit{118.0}     & \textit{62.5}   & 59.5  & \textit{64.9} & \textit{\textbf{42.0}}  & \textit{\textbf{49.5}}  & \textit{63.8}   \\		
		\textbf{Ours}  & \textit{54.0} &  \textit{49.5} &   \textit{43.1} &   \textit{53.0} & \textit{40.4} &  \textit{55.1} &  \textit{54.1} & \textit{57.2} &        \textit{40.1} &    \textit{54.5} &  \textit{39.0} &   \textit{48.7} & \textit{45.7} &   \textit{46.1} &  \textit{53.2} &  \textit{47.7} \\
		\textbf{Ours+refined}  &   \textit{52.7} &  \textit{\textbf{45.4}} &    \textit{\textbf{35.8}} &   \textit{\textbf{51.4}} &     \textit{\textbf{38.8}} &  \textit{\textbf{\textbf{53.1}}} &   \textit{\textbf{53.6}} &  \textit{\textbf{48.0}} &        \textit{\textbf{29.0}} &   \textit{\textbf{43.2}} &  \textit{\textbf{35.6}} & \textit{\textbf{47.2}} & \textit{\textbf{43.0}} &   \textit{45.1} & \textit{50.3} &  \textit{\textbf{43.6}} \\
		\hline
	\end{tabular}
	\caption{The result of mean estimation errors (mm) on Human3.6M.}
	\label{table_human36m2}
\end{table}

In Table~\ref{table_human36m1} and Table~\ref{table_human36m2}, we observed that our approach consistently performs better in categories with more deformations, such as ``Buy", ``Sit", and ``SitDown", etc. For example, on the motion of ``Sit", the best performance is always observed on the results from our approach. Especially, the the-state-of-arts estimation error on ``Sit" is achieved by the proposed approach. However, some exceptions are also observed on categories, including ``Direct.", ``Walk", and ``WalkPair". In these cases, our approach leads the slightly worse performance. This may be due to that the over-fitting issue happens in the proposed approach since the testing poses included in these categories have less deformations.

\subsection{Quantitative Evaluation on HumanEva-I Dataset}\label{s5-3}

The comparative evaluation is also presented in HumanEva-I dataset. Similar to the evaluation of Human3.6M dataset, the quantitative results from both supervised and unsupervised learning approaches are considered, as reported in Table~\ref{table_huamneva}.
\begin{table}[H]\scriptsize
	\renewcommand\arraystretch{0.7}	
	\setlength{\tabcolsep}{1mm}
	\centering
	\begin{tabular}{lrrrrrrr}
		\hline
		\multirow{2}{*}{Method} & \multicolumn{3}{c}{Walking(C1)}                                     & \multicolumn{3}{c}{Jogging(C1)}                           \\
		& S1            & S2            & S3          & S1             & S2            & S3            & Avg        \\
		\hline			
		\multicolumn{8}{c}{\textbf{The supervised learning approach}}    \\  
		\hline
		Simo-Serra et al.~\cite{Simo2013A}       & \textit{65.1}          & \textit{48.6}    & \textit{73.5}     & \textit{74.2}           & \textit{46.6}          & \textit{32.2}          & \textit{56.7}     \\
		Kostrikov et al.~\cite{Kostrikov2014Depth}        & \textit{44.0}   & \textit{30.9}  & \textit{41.7}   & \textit{57.2}           & \textit{35.0}            & \textit{33.3}   & \textit{40.3}           \\
		Bo et al.~\cite{Bo2010Twin}               & \textit{46.4}    & \textit{30.3}   & \textit{64.9}          & 64.5             & \textit{48.0}& \textit{38.2}          & \textit{48.7}            \\				
		Katircioglu et al.~\cite{Katircioglu2018Learning}      & \textit{29.3}  & \textit{17.9}         & \textit{59.5}          &     -    &   -    &   -    & \textit{35.6}     \\	
		Martinez et al.~\cite{martinez2017simple}     & \textit{19.7}  & \textit{17.4} & \textit{46.8}   & \textit{\textbf{26.9}}           & \textbf{18.2}  & \textbf{18.6}          &  \textit{24.6}    \\
		Pavlakos et al.~\cite{pavlakos2017volumetric}     & \textit{22.3}  & \textit{19.5} & \textit{29.7}     & \textit{28.9}           & \textit{21.9}  & \textit{23.8}     &       \textit{24.3}    \\				
		Morenonoguer et al.~\cite{Morenonoguer20173D}     & \textbf{19.7}  & \textbf{13.0} & \textit{\textbf{24.9}} & \textit{39.7}   & \textit{20.0}  & \textit{21.0}       & \textit{23.1}   \\
		\hline
		\multicolumn{8}{c}{\textbf{The unsupervised learning approach}}  \\  
		\hline
		Radwan et al.~\cite{Radwan2013Monocular}        & \textit{75.1}          & \textit{99.8}          & \textit{93.8}            & \textit{79.2}           & \textit{89.8}          & \textit{99.4}          & \textit{89.5}          \\	
		Wang et al.~\cite{Wang2014Robust}             & \textit{71.9}    & \textit{75.7}    & \textit{85.3}         & \textit{62.6}          &\textit{77.7}          & \textit{54.4}          & \textit{71.3}           \\
		Yasin et al.~\cite{Yasin2016A}   & \textit{35.8}          & \textit{32.4}         & \textit{41.6}        & \textit{46.6}          & \textit{41.4}          & \textit{35.4}          & \textit{38.9}           \\					
		Wang et al.~\cite{Wang2018Robust}     &\textit{40.3}  & \textit{37.6} & \textit{37.4}         & \textit{39.7}        & 36.2  & \textit{38.4}          &    \textit{38.3} \\	
		Zhou et al.~\cite{Zhou2017MonoCap}   & \textit{34.3}     & \textit{31.6}   & \textit{49.3}    & \textit{48.6}         & \textit{34.0}     & \textit{30.0}  & \textit{37.9}      \\						
		\textbf{Ours}   & \textit{34.5}     & \textit{30.7} & \textit{32.8}    &\textit{\textbf{32.3}}   & \textit{33.7}   &\textit{32.4}     &\textit{32.7}   \\
		\textbf{Ours+refined}   & \textit{\textbf{29.4}}     & \textit{\textbf{29.2}}  & \textit{\textbf{30.2}}    & \textit{32.7}    & \textit{\textbf{32.2}}      &\textit{\textbf{29.9}}      &\textit{\textbf{30.6}}    \\
		\hline			
	\end{tabular}
	\caption{The result of mean estimation errors (mm) on HumanEva-I dataset.}
	\label{table_huamneva}
\end{table}
We clearly observed that our approach outperforms all unsupervised learning approaches by more than \textit{13}\% on average. This figure is further increased to \textit{19}\% when we use better initialization. In addition, even compared to supervised learning approaches, the performance of our approach is still comparable. Specially, the proposed approach outperforms some competitive works \cite{Simo2013A, Kostrikov2014Depth, Bo2010Twin, Katircioglu2018Learning} by more than \textit{8}\% on average.

Although our approach without refined initialization outperforms other unsupervised learning methods in most categories, there are still a few exceptions. For example, the performances of our approach are slightly worse than the literature \cite{Zhou2017MonoCap} on some categories, such as ``Walking" of the subject the ``S1" and ``Jogging" of the subject ``S3". However, the accuracies on these cases are considerably improved when we use a better initialization. In addition, notice that the work \cite{Zhou2017MonoCap} action specific dictionaries for each subject separately. However, for the model generality, we train our dictionaries without distinction of specific action or subjects.
 
In most cases, a fine initialization usually boosts the performance of our approach. However, there are also very few exceptions. As we observed in Table~\ref{table_huamneva}, the fine initialization (\textbf{Ours+refined}) leads to a slightly worse reconstruction in ``Jogging" of subject ``S1". It may happen that the solution provided by the convex approach is not a good initialization.

\subsection{Quantitative Evaluation on CMU MoCap Dataset} \label{s5-4}
Compared to comprehensive evaluations on HumanEva-I and Human3.6m, there are fewer approaches to evaluate on CMU Mocap. However, we notice that still some generalization experiments in this dataset are reported on some literature~\cite{Ramakrishna2012Reconstructing, Zhou2016Sparse} which both are SR based approaches. Since these approaches are the most related works to our approach, we also evaluate on this dataset to show the effectiveness of the proposed approach.

\begin{figure}[H]
	\centering	
	\subfloat[]{\label{Fig:CMU2}
		\includegraphics[scale=0.37]{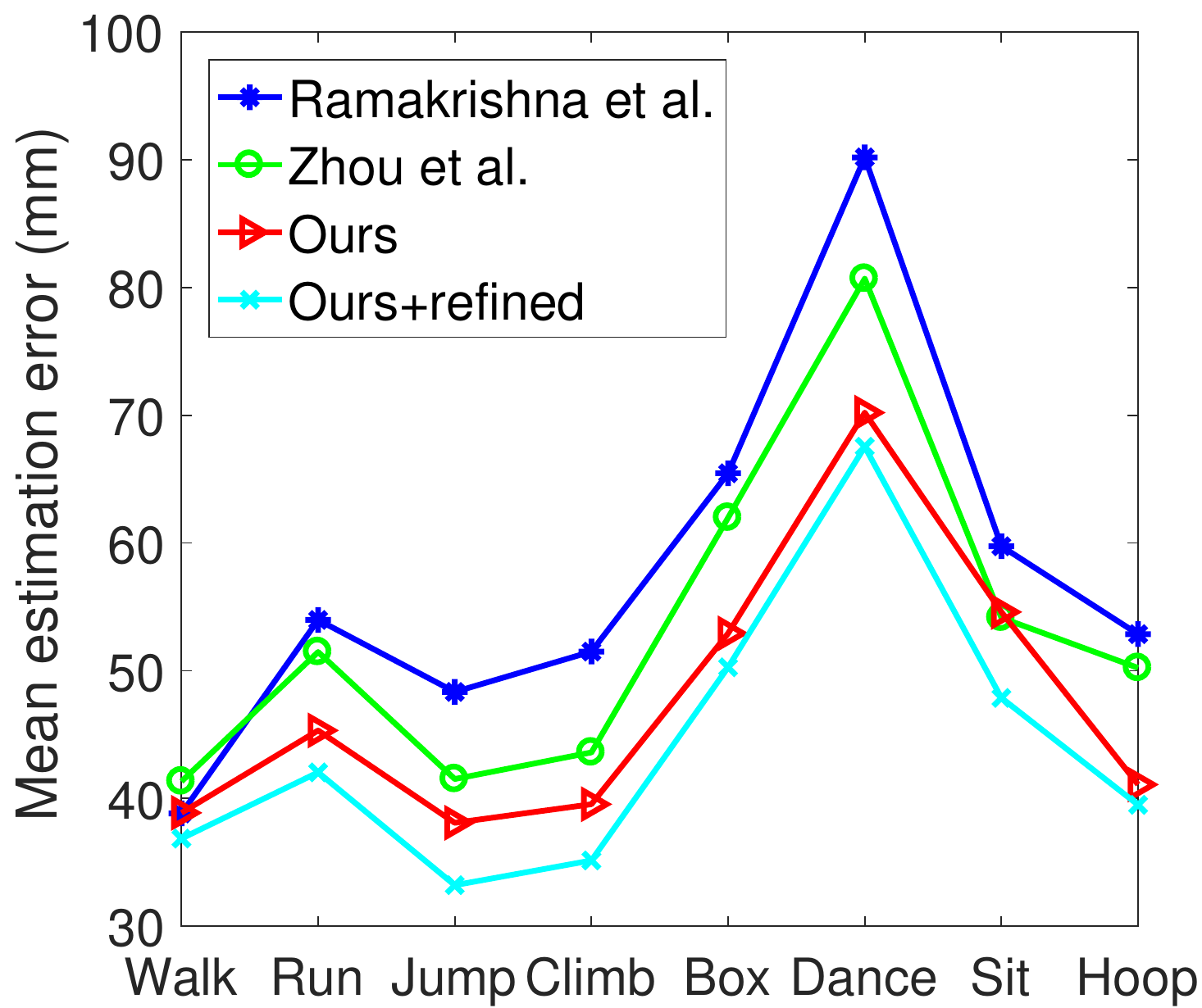}} 
	\subfloat[]{\label{Fig:CMU3}
		\includegraphics[scale=0.37]{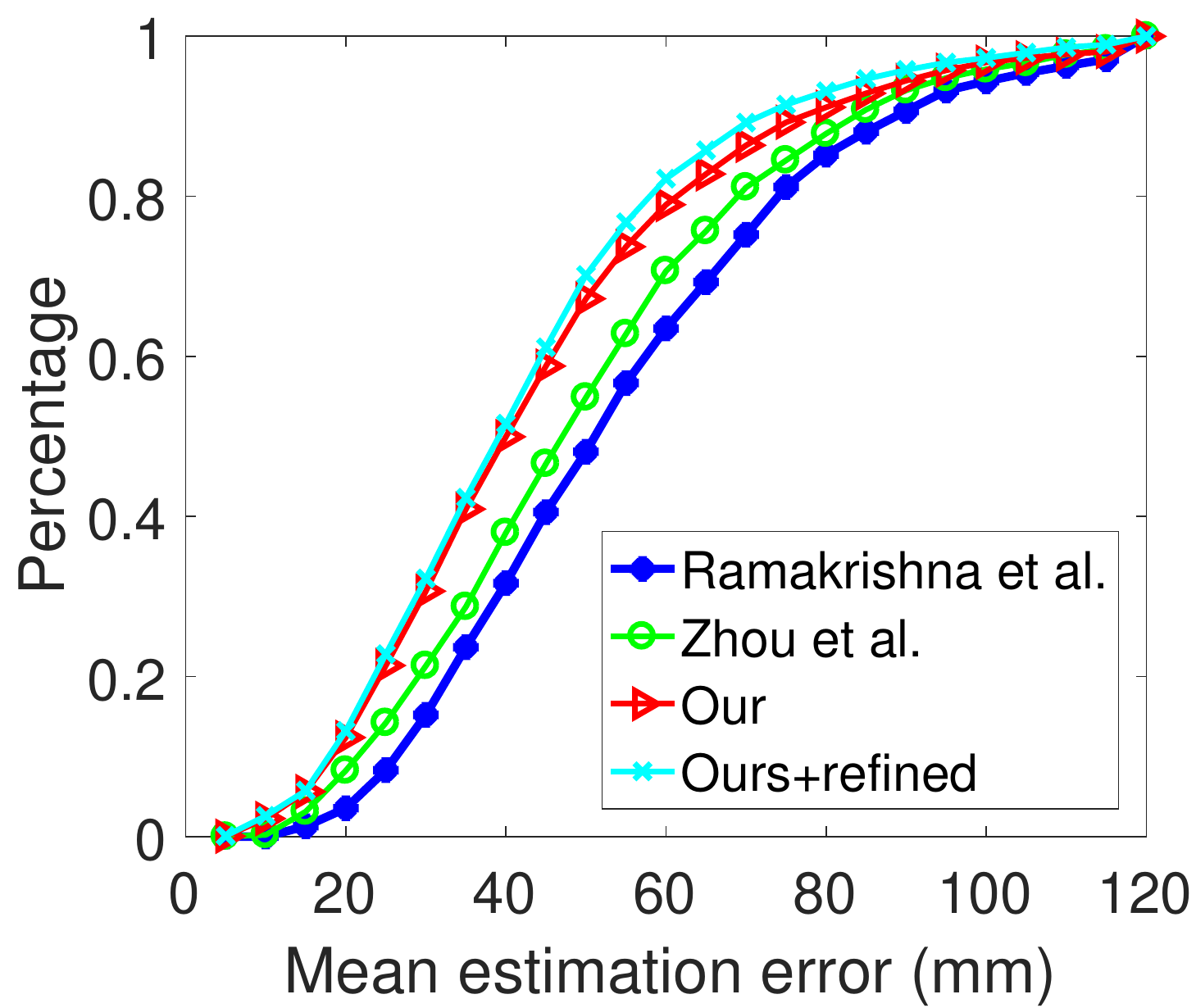}} 
	\subfloat[]{\label{Fig:CMU4}
		\includegraphics[scale=0.37]{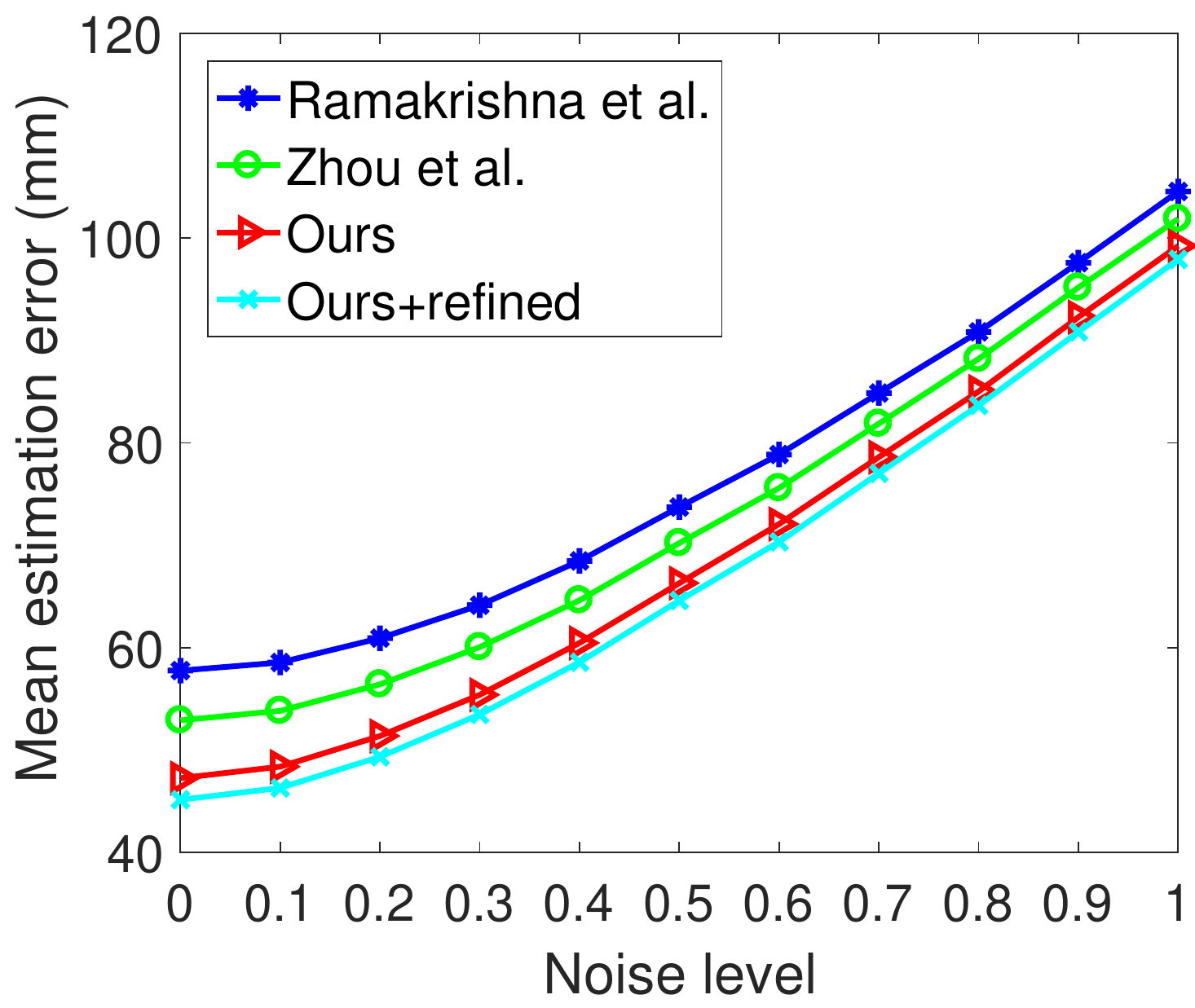}}
	\caption{The quantitative results on CMU MoCap. (a) The mean estimation errors of different motions. (b) Distribution of the mean estimation for all testing examples. The y-axis is the percentage of cases where the estimation error is less than the corresponding x-axis value. (c) The sensitivity to Gaussian noise. }
	\label{Fig:RecAccuracy_CMU}		
\end{figure}

To evaluate our model on different motions, we observed that the proposed approaches outperforms all compared baselines across all motions categories, as shown in Figure \ref{Fig:RecAccuracy_CMU}(a). Due to the smaller motion range and deformation of ``Walk" category, all the compared algorithms have lower errors and no evident differences between them. However, in some more complex categories, such as ``Dance" and ``Hoop", we can clearly see that our algorithms boost performance. In addition, we present the percentage of different error ranges, as shown in Figure \ref{Fig:RecAccuracy_CMU}(b). The proposed approaches perform better reconstruction on most testing examples than the baselines. Specially, the percentage of our approach (\textbf{Ours+refined}) exceeds \textit{80\%} when the mean estimation errors are smaller than \textit{60} (mm). However, this percentage is decreased to about \textit{60\%} achieved by the~\cite{Ramakrishna2012Reconstructing} and about \textit{70\%} by~\cite{Zhou2016Sparse}. Finally, we evaluate the robustness of our approach against the Gaussian noises with various standard deviations, as shown in \ref{Fig:RecAccuracy_CMU}(c). The accuracies of our approach are consistently better than the compared approaches under all noise levels.

In addition, the mean estimation errors on different \textit{15} body joints, including Head, Neck, Left Shoulder (LS), Right Shoulder (RS), etc., are shown in Table \ref{table_cmu}.
\begin{table}[H]\scriptsize
	\renewcommand\arraystretch{0.7}	
	\setlength{\tabcolsep}{1mm}
	\centering
	\begin{tabular}{lrrrrrrrrrrrrrrrr}
		\hline
		Methods	& Pelvis & RH & RK & RA & LH & LK & LA & Neck & Head & PS & RE & RW & LS & LE & LW & Avg \\
		\hline 
		Ramakrishna et al.~\cite{Ramakrishna2012Reconstructing}	& \textit{40.0} &  \textit{48.4}  &\textit{59.8}  &\textit{69.5}  &\textit{48.2} & \textit{61.9} & \textit{74.9} & \textit{29.0} & \textit{46.6} & \textit{41.6} & \textit{56.6} &\textit{89.8} & \textit{46.8} & \textit{64.3}& \textit{88.8} &\textit{57.8} \\
		Zhou et al.~\cite{Zhou2016Sparse}	& \textit{35.3} &\textit{45.5} &\textit{53.3}  &\textit{64.2} &\textit{48.3}  &\textit{53.8}  &\textit{68.1} &\textit{25.6} &\textit{44.5} &\textit{39.9} &\textit{58.0} &\textit{76.7} &\textit{43.7} &\textit{61.0} &\textit{75.7} &\textit{52.9} \\
		\textbf{Ours}	&\textit{33.1} & \textit{42.8} &\textit{49.2} &\textit{59.1} &\textit{41.3} &\textit{48.8} &\textit{67.4} &\textit{23.2} &\textit{36.2} &\textit{32.7} &\textit{51.3} &\textit{68.7} &\textit{37.4} &\textit{49.3} &\textit{68.4} & \textit{47.3}\\
		\textbf{Ours+refined}& \textit{\textbf{30.5}}& \textit{\textbf{39.2}}&  \textit{\textbf{45.9}}& \textit{\textbf{57.7}}& \textbf{38.4}& \textit{\textbf{47.5}}& \textit{\textbf{64.7}}& \textit{\textbf{23.1}}& \textit{\textbf{34.6}}& \textit{\textbf{30.9}}& \textit{\textbf{47.9}}& \textit{\textbf{68.7}}& \textit{\textbf{35.8}}& \textit{\textbf{46.8}}& \textit{\textbf{65.3}} &\textit{\textbf{45.1}} \\
		\hline  
	\end{tabular}
	\caption{The mean estimation errors (mm) of different joints on CMU MoCap}
	\label{table_cmu}
\end{table}
As seen from the table, our approaches consistently achieve lower estimation errors across all body joints than compared baselines. Specially, more than \textit{10\%} (\textbf{Ours}) and \textit{14\%} (\textbf{Ours+refined}) on average improvements are achieved by the proposed approaches. Since severe deformation and self-occlusion are more likely to occur in some joints, such as left wrist (LW), right wrist (RW), left ankle (LA), and right ankle (RA), the performances of all compared algorithms are usually worse on these joints. However, the proposed approach shows more evident improvements in these joints. 

\subsection{Qualitative Evaluation on MPII Dataset}\label{s5-5}
\begin{figure}[ht]
	\centering
	\includegraphics[scale=0.49]{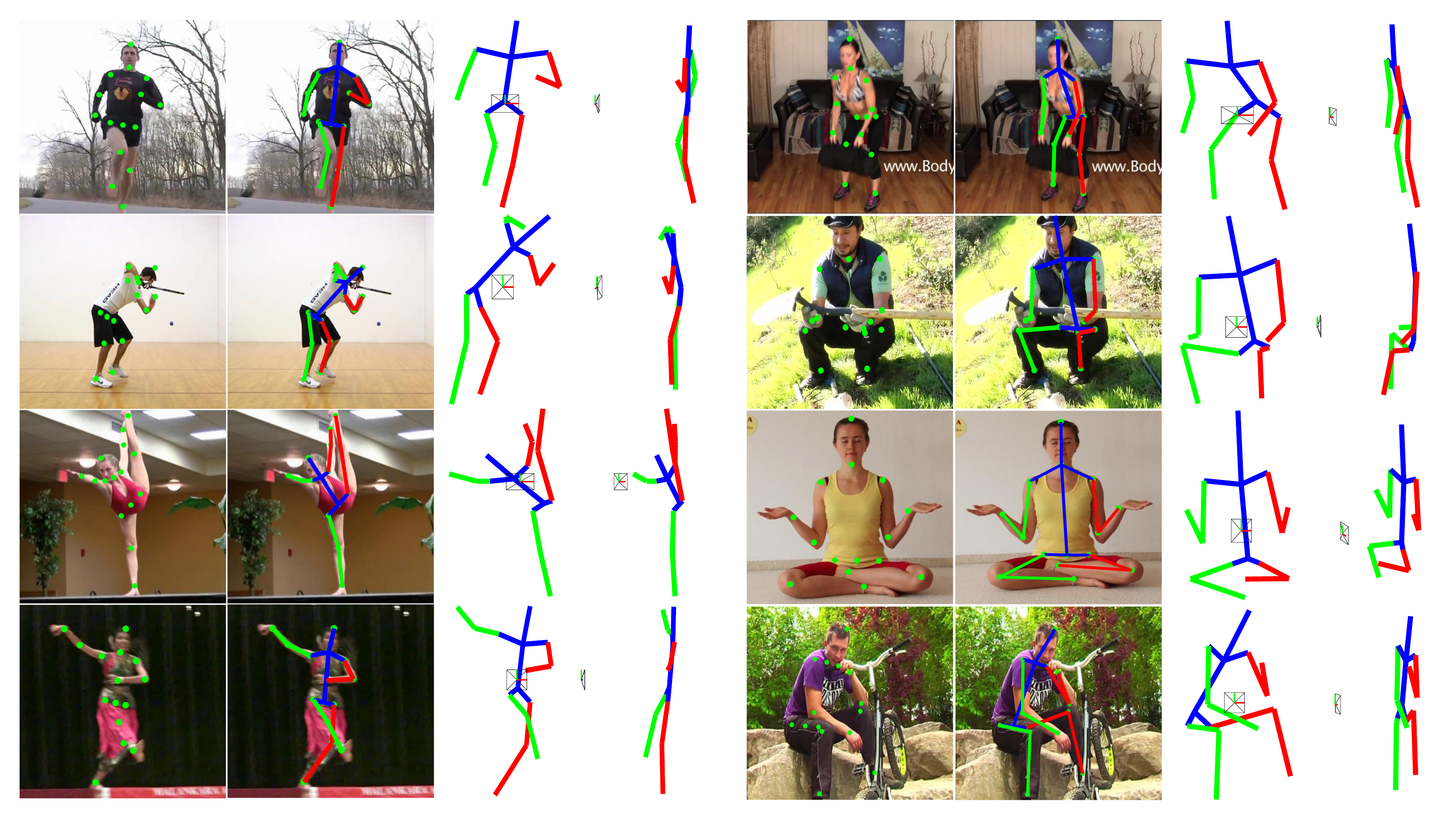}
	\caption{Example results on MPII. Two examples are included in each row. In each example, the figures from left to right correspond to the original image with 2D joints, the ground-truth 2D pose, and the estimated 3D pose by the proposed approach. Notice that the estimated 3D pose is shown in two novel views.}
	\label{fig:mpiiexample}
\end{figure}

\begin{figure}[H]
	\centering
	\includegraphics[scale=0.49]{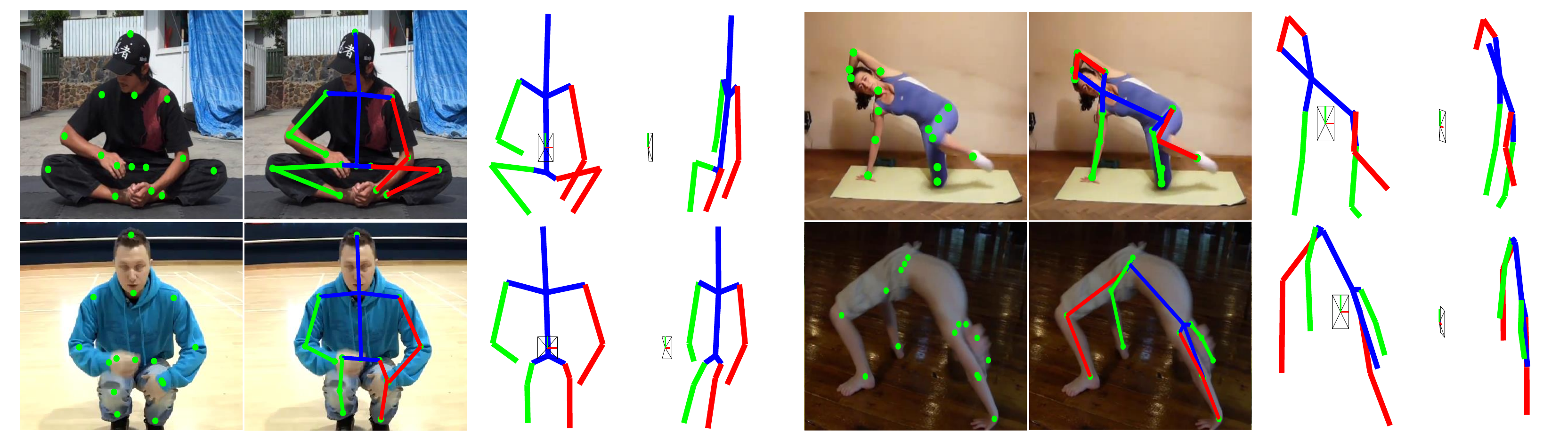}
	\caption{Failure examples on MPII. Two examples are included in each row. In each example, the figures from left to right correspond to the original image with 2D joints, the ground-truth 2D pose, and the estimated 3D pose by the proposed approach. Notice that the estimated 3D pose is shown in two novel views.}
	\label{fig:fmpiiexample}
\end{figure}

The applicability of our approach for the 3D pose estimation on in-the-wild images is also demonstrated. To evaluate the effectiveness of the proposed approach, we selected eight various actions and challenging examples. The qualitative results are presented in Figure~\ref{fig:mpiiexample}. 
Notice that this dataset did not provide 3D annotations, thus we use the dictionary learned from CMU Mocap dataset for the 3D pose inference. Even so, we clearly observed that the proposed method recovers the 3D human poses successfully from various actions and subjects. However, we also represent some failure cases, as shown in Figure~\ref{fig:fmpiiexample}. Similar to most algorithms, the depth ambiguity and serious occlusions are still problems.

\section{Conclusion}\label{s6}
In this work, we aim to enhance the standard SR model to recover various complex poses under available limited training data. Firstly, we proposed the Shape Decomposition Model (SDM) in which a complex 3D pose is represented as two parts which are the global structure and corresponding deformations. 
Considering that distributions between the global structure and deformations of the 3D poses are different, we represented these two parts as the sparse and dense linear combination of 3D bases respectively. In addition, we proposed dual dictionary learning approach to learn two sets of 3D bases for representations of the 3D pose. Our dictionary learning model captures more implicit information as priors from limited training data. The effectiveness of our approach has been demonstrated on several evaluation benchmarks. In comparison, the proposed approach shown superior performance across all quantitative studies than all compared unsupervised learning approaches and even many competitive supervised methods. Especially, our approach shown more significant improvements in some categories with more deformations. For example, on the category "Sit" of Human3.6M, our approach achieved state-of-the-art performance on estimation error metric with a wide margin (more than \textit{66\%} and \textit{28\%} improvements respectively comparing to all considered unsupervised and supervised learning approaches). In addition, the qualitative evaluation on in-the-wild images shown that our approach is able to estimate satisfactory 3D pose even without 3D annotation supervision. Future, we may extend our model to other non-rigid 3D object estimations, such as 3D face reconstruction, 3D hand pose estimation, etc.  

\section*{APPENDIX A} \label{APPENDIXA}
\subsection*{The ALGORITHM TO SOLVE JOINT DICTIONARY LEARNING}
The algorithm to solve Eq.~\eqref{4-4} is presented here. For each training pose ${D_m}$, where $\forall m \in [1,N]$, we rewritten the cost function as 
\begin{equation}\label{eqe:apendix}\small
\begin{aligned} 
\mathcal{L}(\bm{B_u^*},\bm{B_v^*},\bm{C_u},\bm{C_v})= \frac{1}{2} \Vert \bm{D}_m-\sum_{j=1}^k (\bm{B}_{\bm{u}j}c_{ujm}+\bm{B}_{\bm{v}j}c_{vjm})\Vert_2^2 
+ \gamma\Vert \bm{C_u} \Vert_1 + \eta\Vert \bm{C_v} \Vert_2^2,
\end{aligned}
\end{equation}
where $\bm{B_u^*}$ and $\bm{B_v^*}$ are the concatenation of $\bm{B}_{uj}$ and $\bm{B}_{vj}$ respectively. The problems in \eqref{eqe:apendix} are solved by alternately updating strategy via projected gradient descent. The joint dictionary learning algorithm is summarized in Algorithm \ref{alg:diclearn}.
\begin{algorithm}[H]\small
	\caption{Joint Dictionary Learning Algorithm.}
	\label{alg:diclearn}
	\textbf{Input}: $\bm{D}_1,...,\bm{D}_m$ \quad \textcolor[RGB]{88,87,86}{//The training examples} \\
	\textbf{Parameter}: $\tau, \gamma, \eta, \delta_1, \delta_2, \delta_3, \delta_4, k$ \textcolor[RGB]{88,87,86}{//$\tau$ is a tolerance for object value. $\gamma, \eta$ are balance factors. $\delta_{1,2,3,4}$ are step sizes. $k$ is the dictionary size} \\
	\textbf{Output}: $\{\bm{B}_{\bm{u}j}\}_{j=1}^k,、\{\bm{B}_{\bm{v}j}\}_{j=1}^k$
	\begin{algorithmic}[1] 
		\STATE initialize $\bm{C_u}, \bm{C_v}, \bm{B_u^*}, \bm{B_v^*}$ and parameters;
		\WHILE{$\Vert\bm{r}\Vert_2>\tau$ or $\ell<\ell_{max}$}
		\STATE $\bm{C_u} \leftarrow \delta_1\bigtriangledown_{\bm{C_u}}\mathcal{L}(\bm{B_u^*},\bm{B_v^*},\bm{C_u},\bm{C_v})$;
		\STATE $\bm{C_v} \leftarrow \delta_2\bigtriangledown_{\bm{C_v}}\mathcal{L}(\bm{B_u^*},\bm{B_v^*},\bm{C_u},\bm{C_v})$;
		\STATE $\bm{B_u^*} \leftarrow \delta_3\bigtriangledown_{\bm{B_u^*}}\mathcal{L}(\bm{B_u^*},\bm{B_v^*},\bm{C_u},\bm{C_v})$;
		\STATE $\bm{B_v^*} \leftarrow \delta_4\bigtriangledown_{\bm{B_v^*}}\mathcal{L}(\bm{B_u^*},\bm{B_v^*},\bm{C_u},\bm{C_v})$;
		\STATE \emph{\textbf{update}} $\ell=\ell+1$; \quad \textcolor[RGB]{88,87,86} {// iteration count.} 
		\ENDWHILE
	\end{algorithmic}
\end{algorithm}

\section*{Acknowledgment}
This research was supported by the National Nature Science Foundation of China (grant no 61671397).

\section*{References}\label{s8}
\bibliography{mybibfile}

\begin{thebibliography}{10}
\expandafter\ifx\csname url\endcsname\relax
  \def\url#1{\texttt{#1}}\fi
\expandafter\ifx\csname urlprefix\endcsname\relax\def\urlprefix{URL }\fi
\expandafter\ifx\csname href\endcsname\relax
  \def\href#1#2{#2} \def\path#1{#1}\fi

\bibitem{Jung2015Random}
H.~Y. Jung, S.~Lee, S.~H. Yong, I.~D. Yun, Random tree walk toward
  instantaneous 3{D} human pose estimation, in: Computer Vision and Pattern
  Recognition, 2015, pp. 2467--2474.

\bibitem{rhodin2018learning}
H.~Rhodin, J.~Sp{\"o}rri, I.~Katircioglu, V.~Constantin, F.~Meyer,
  E.~M{\"u}ller, M.~Salzmann, P.~Fua, Learning monocular 3{D} human pose
  estimation from multi-view images, in: Computer Vision and Pattern
  Recognition, 2018, pp. 8437--8446.

\bibitem{Shen2013Part}
J.~Shen, W.~Yang, Q.~Liao, Part template: 3{D} representation for multiview
  human pose estimation, Pattern Recognition 46~(7) (2013) 1920--1932.

\bibitem{Bo2010Twin}
L.~Bo, C.~Sminchisescu, Twin gaussian processes for structured prediction,
  International Journal of Computer Vision 87~(1-2) (2010) 28--52.

\bibitem{Jiang20103D}
H.~Jiang, 3{D} human pose reconstruction using millions of exemplars, in:
  International Conference on Pattern Recognition, 2010, pp. 1674--1677.

\bibitem{Chen20113D}
C.~Chen, Y.~Yang, F.~Nie, J.~M. Odobez, 3{D} human pose recovery from image by
  efficient visual feature selection, Computer Vision and Image Understanding
  115~(3) (2011) 290--299.

\bibitem{Li20143D}
S.~Li, A.~B. Chan, 3{D} human pose estimation from monocular images with deep
  convolutional neural network, in: Asian Conference on Computer Vision, 2014,
  pp. 332--347.

\bibitem{Tekin2016Structured}
B.~Tekin, I.~Katircioglu, M.~Salzmann, V.~Lepetit, P.~Fua, Structured
  prediction of 3{D} human pose with deep neural networks, arXiv preprint
  arXiv:1605.05180.

\bibitem{Katircioglu2018Learning}
I.~Katircioglu, B.~Tekin, M.~Salzmann, V.~Lepetit, P.~Fua, Learning latent
  representations of 3{D} human pose with deep neural networks, International
  Journal of Computer Vision 126~(12) (2018) 1--16.

\bibitem{Ramakrishna2012Reconstructing}
V.~Ramakrishna, T.~Kanade, Y.~Sheikh, Reconstructing 3{D} human pose from 2{D}
  image landmarks, in: European Conference on Computer Vision, 2012, pp.
  573--586.

\bibitem{Zhou2016Sparse}
X.~Zhou, M.~Zhu, S.~Leonardos, K.~Daniilidis, Sparse representation for 3{D}
  shape estimation: A convex relaxation approach, IEEE Transactions on Pattern
  Analysis and Machine Intelligence 39~(8) (2017) 1648--1661.

\bibitem{Zhou2016Sparseness}
X.~Zhou, M.~Zhu, S.~Leonardos, K.~G. Derpanis, K.~Daniilidis, Sparseness meets
  deepness: 3{D} human pose estimation from monocular video, in: Computer
  Vision and Pattern Recognition, 2016, pp. 4966--4975.

\bibitem{Bogo2016Keep}
F.~Bogo, A.~Kanazawa, C.~Lassner, P.~Gehler, J.~Romero, M.~J. Black, Keep it
  smpl: Automatic estimation of 3{D} human pose and shape from a single image,
  in: European Conference on Computer Vision, 2016, pp. 561--578.

\bibitem{Newell2016Stacked}
A.~Newell, K.~Yang, J.~Deng, Stacked hourglass networks for human pose
  estimation, in: European Conference on Computer Vision, 2016, pp. 483--499.

\bibitem{Olshausen1997Sparse}
B.~A. Olshausen, D.~J. Field, Sparse coding with an overcomplete basis set: a
  strategy employed by v1?, Vision Research 37~(23) (1997) 3311--3325.

\bibitem{CMU}
Mocap: Carnegie mellon university motion capture database,
  \url{http://mocap.cs.cmu.edu/}.

\bibitem{Sigal2006HumanEva}
L.~Sigal, M.~J. Black, Humaneva: Synchronized video and motion capture dataset
  for evaluation of articulated human motion, International Journal of Computer
  Vision 87~(1-2) (2006) 4--27.

\bibitem{Ionescu2014Human3}
C.~Ionescu, D.~Papava, V.~Olaru, C.~Sminchisescu, Human3.6m: Large scale
  datasets and predictive methods for 3{D} human sensing in natural
  environments, IEEE Transactions on Pattern Analysis and Machine Intelligence
  36~(7) (2014) 1325--1339.

\bibitem{Zhou2017MonoCap}
X.~Zhou, M.~Zhu, G.~Pavlakos, S.~Leonardos, K.~G. Derpanis, K.~Daniilidis,
  Monocap: Monocular human motion capture using a cnn coupled with a geometric
  prior, IEEE Transactions on Pattern Analysis and Machine Intelligence 41~(4)
  (2019) 901--914.

\bibitem{Wang2018Robust}
C.~Wang, Y.~Wang, Z.~Lin, A.~Yuille, Robust 3{D} human pose estimation from
  single images or video sequences, IEEE Transactions on Pattern Analysis and
  Machine Intelligence 41~(5) (2019) 1227--1241.

\bibitem{dou2018monocular}
P.~Dou, Y.~Wu, S.~K. Shah, I.~A. Kakadiaris, Monocular 3{D} facial shape
  reconstruction from a single 2{D} image with coupled-dictionary learning and
  sparse coding, Pattern Recognition 81 (2018) 515--527.

\bibitem{Xu2017Recovering}
C.~Xu, Z.~Zhang, B.~Wang, G.~Hu, E.~R. Hancock, Recovering variations in facial
  albedo from low resolution images, Pattern Recognition 74 (2017) 373--384.

\bibitem{Shao2018Spatial}
Y.~Shao, S.~Nong, C.~Gao, M.~Li, Spatial and class structure regularized sparse
  representation graph for semi-supervised hyperspectral image classification,
  Pattern Recognition 81 (2018) 102--114.

\bibitem{Chen2016Synthesizing}
W.~Chen, H.~Wang, Y.~Li, H.~Su, Z.~Wang, C.~Tu, D.~Lischinski, D.~Cohen-Or,
  B.~Chen, Synthesizing training images for boosting human 3{D} pose
  estimation, in: International Conference on 3d Vision, 2016, pp. 479--488.

\bibitem{elgammal2004inferring}
A.~Elgammal, C.-S. Lee, Inferring 3{D} body pose from silhouettes using
  activity manifold learning, in: Computer Vision and Pattern Recognition,
  2004, pp. 681--688.

\bibitem{Agarwal20043D}
A.~Agarwal, B.~Triggs, 3{D} human pose from silhouettes by relevance vector
  regression, in: Computer Vision and Pattern Recognition, 2004, pp. 882--888.

\bibitem{Sedai2013Discriminative}
S.~Sedai, M.~Bennamoun, D.~Q. Huynh, Discriminative fusion of shape and
  appearance features for human pose estimation, Pattern Recognition 46~(12)
  (2013) 3223--3237.

\bibitem{Toshev2013DeepPose}
A.~Toshev, C.~Szegedy, Deeppose: Human pose estimation via deep neural
  networks, in: Computer Vision and Pattern Recognition, 2013, pp. 1653--1660.

\bibitem{Li2016Maximum}
S.~Li, W.~Zhang, A.~B. Chan, Maximum-margin structured learning with deep
  networks for 3{D} human pose estimation, International Journal of Computer
  Vision 122~(1) (2017) 149--168.

\bibitem{Morenonoguer20173D}
F.~Morenonoguer, 3{D} human pose estimation from a single image via distance
  matrix regression, in: Computer Vision and Pattern Recognition, 2017, pp.
  1561--1570.

\bibitem{pavlakos2017volumetric}
G.~Pavlakos, X.~Zhou, K.~G. Derpanis, K.~Daniilidis, Coarse-to-fine volumetric
  prediction for single-image 3{D} human pose, in: Computer Vision and Pattern
  Recognition, 2017, pp. 1263--1272.

\bibitem{martinez2017simple}
J.~Martinez, R.~Hossain, J.~Romero, J.~J. Little, A simple yet effective
  baseline for 3{D} human pose estimation, in: International Conference on
  Computer Vision, 2017, pp. 2659--2668.

\bibitem{pavlakos2018ordinal}
G.~Pavlakos, X.~Zhou, K.~Daniilidis, Ordinal depth supervision for 3{D} human
  pose estimation, in: Computer Vision and Pattern Recognition, 2018, pp.
  7307--7316.

\bibitem{varol2017learning}
G.~Varol, J.~Romero, X.~Martin, N.~Mahmood, M.~Black, I.~Laptev, C.~Schmid,
  Learning from synthetic humans, in: Computer Vision and Pattern Recognition,
  2017, pp. 4627--4635.

\bibitem{Simo2012Single}
E.~Simo-Serra, A.~Ramisa, G.~Alenyà, C.~Torras, Single image 3{D} human pose
  estimation from noisy observations, in: Computer Vision and Pattern
  Recognition, 2012, pp. 2673--2680.

\bibitem{Simo2013A}
E.~Simo-Serra, A.~Quattoni, C.~Torras, F.~Moreno-Noguer, A joint model for 2{D}
  and 3{D} pose estimation from a single image, in: Computer Vision and Pattern
  Recognition, 2013, pp. 3634--3641.

\bibitem{Sanzari2016Bayesian}
M.~Sanzari, V.~Ntouskos, F.~Pirri, Bayesian image based 3{D} pose estimation,
  in: European Conference on Computer Vision, 2016, pp. 566--582.

\bibitem{Chen20163D}
C.~H. Chen, D.~Ramanan, 3{D} human pose estimation = 2{D} pose estimation +
  matching, in: Computer Vision and Pattern Recognition, 2017, pp. 5759--5767.

\bibitem{Cootes1995Active}
T.~F. Cootes, C.~J. Taylor, D.~H. Cooper, J.~Graham, Active shape models-their
  training and application, Computer Vision and Image Understanding 61~(1)
  (1995) 38--59.

\bibitem{Kostrikov2014Depth}
I.~Kostrikov, J.~Gall, Depth sweep regression forests for estimating 3{D} human
  pose from images, in: British Machine Vision Conference, 2014, pp. 1--13.

\bibitem{Radwan2014Monocular}
I.~Radwan, A.~Dhall, R.~Goecke, Monocular image 3{D} human pose estimation
  under self-occlusion, in: IEEE International Conference on Computer Vision,
  2014, pp. 1888--1895.

\bibitem{Yu2016Marker}
Y.~Du, Y.~Wong, Y.~Liu, F.~Han, Y.~Gui, Z.~Wang, M.~Kankanhalli, W.~Geng,
  Marker-less 3{D} human motion capture with monocular image sequence and
  height-maps, in: European Conference on Computer Vision, 2016, pp. 20--36.

\bibitem{Sarafianos20163D}
N.~Sarafianos, B.~Boteanu, B.~Ionescu, I.~A. Kakadiaris, 3{D} human pose
  estimation: A review of the literature and analysis of covariates, Computer
  Vision and Image Understanding 152 (2016) 1--20.

\bibitem{Candes2006Near}
E.~J. Candes, T.~Tao, Near-optimal signal recovery from random projections:
  Universal encoding strategies?, IEEE Transactions on Information Theory
  52~(12) (2006) 5406--5425.

\bibitem{Wang2014Robust}
C.~Wang, Y.~Wang, Z.~Lin, A.~L. Yuille, W.~Gao, Robust estimation of 3{D} human
  poses from a single image, in: Computer Vision and Pattern Recognition, 2014,
  pp. 2369--2376.

\bibitem{Xudong2015Sparse}
J.~Xudong, L.~Jian, Sparse and dense hybrid representation via dictionary
  decomposition for face recognition, IEEE Transactions on Pattern Analysis and
  Machine Intelligence 37~(5) (2015) 1067--1079.

\bibitem{boyd2011distributed}
S.~P. Boyd, N.~Parikh, E.~Chu, B.~Peleato, J.~Eckstein, Distributed
  optimization and statistical learning via the alternating direction method of
  multipliers, Foundations and Trends in Machine Learning archive 3~(1) (2011)
  1--122.

\bibitem{Boumal2013Manopt}
N.~Boumal, B.~Mishra, P.~A. Absil, R.~Sepulchre, Manopt, a matlab toolbox for
  optimization on manifolds, Journal of Machine Learning Research 15~(1) (2013)
  1455--1459.

\bibitem{Nesterov2013Gradient}
Nesterov, Yu, Gradient methods for minimizing composite functions, Mathematical
  Programming 140~(1) (2013) 125--161.

\bibitem{Andriluka20142D}
M.~Andriluka, L.~Pishchulin, P.~Gehler, B.~Schiele, 2{D} human pose estimation:
  New benchmark and state of the art analysis, in: Computer Vision and Pattern
  Recognition, 2014, pp. 3686--3693.

\bibitem{Tekin2016Direct}
B.~Tekin, A.~Rozantsev, V.~Lepetit, P.~Fua, Direct prediction of 3{D} body
  poses from motion compensated sequences, in: Computer Vision and Pattern
  Recognition, 2016, pp. 991--1000.

\bibitem{Zhou2016Deep}
X.~Zhou, X.~Sun, W.~Zhang, S.~Liang, Y.~Wei, Deep kinematic pose regression,
  in: European Conference on Computer Vision, 2016, pp. 186--201.

\bibitem{Radwan2013Monocular}
I.~Radwan, A.~Dhall, R.~Goecke, Monocular image 3{D} human pose estimation
  under self-occlusion, in: International Conference on Computer Vision, 2013,
  pp. 1888--1895.

\bibitem{Yasin2016A}
H.~Yasin, U.~Iqbal, B.~Krüger, A.~Weber, J.~Gall, A dual-source approach for
  3{D} pose estimation from a single image, in: Computer Vision and Pattern
  Recognition, 2016, pp. 4948--4956.

\bibitem{Zhou20153D}
X.~Zhou, S.~Leonardos, X.~Hu, K.~Daniilidis, 3{D} shape estimation from 2d
  landmarks: A convex relaxation approach, in: Computer Vision and Pattern
  Recognition, 2015, pp. 4447--4455.

\bibitem{tekin2017learning}
B.~Tekin, P.~Marquez~Neila, M.~Salzmann, P.~Fua, Learning to fuse 2{D} and 3{D}
  image cues for monocular body pose estimation, in: International Conference
  on Computer Vision, 2017, pp. 3961--3970.

\bibitem{Nie2017Monocular}
B.~X. Nie, P.~Wei, S.~C. Zhu, Monocular 3{D} human pose estimation by
  predicting depth on joints, in: International Conference on Computer Vision,
  2017, pp. 3467--3475.

\bibitem{tome2017lifting}
D.~Tome, C.~Russell, L.~Agapito, Lifting from the deep: Convolutional 3{D} pose
  estimation from a single image, in: Computer Vision and Pattern Recognition,
  2017, pp. 2500--2509.

\bibitem{rogez2017lcr}
G.~Rogez, P.~Weinzaepfel, C.~Schmid, Lcr-net:
  Localization-classification-regression for human pose, in: Computer Vision
  and Pattern Recognition, 2017, pp. 1216--1224.

\bibitem{zhou2017towards}
X.~Zhou, Q.~Huang, X.~Sun, X.~Xue, Y.~Wei, Towards 3{D} human pose estimation
  in the wild: a weakly-supervised approach, in: International Conference on
  Computer Vision, 2017, pp. 398--407.

\bibitem{chen20173d}
C.-H. Chen, D.~Ramanan, 3{D} human pose estimation = 2{D} pose estimation+
  matching, in: Computer Vision and Pattern Recognition, 2017, pp. 5759--5767.

\bibitem{Lin2017Recurrent}
M.~Lin, L.~Liang, X.~Liang, K.~Wang, C.~Hui, M.~Lin, L.~Liang, X.~Liang,
  K.~Wang, C.~Hui, Recurrent 3{D} pose sequence machines, in: Computer Vision
  and Pattern Recognition, 2017, pp. 5543--5552.

\bibitem{Akhter2015Pose}
I.~Akhter, M.~J. Black, Pose-conditioned joint angle limits for 3{D} human pose
  reconstruction, in: Computer Vision and Pattern Recognition, 2015, pp.
  1446--1455.

\bibitem{wang20193d}
K.~Wang, L.~Lin, C.~Jiang, C.~Qian, P.~Wei, 3{D} human pose machines with
  self-supervised learning, IEEE Transactions on Pattern Analysis and Machine
  Intelligence (2019) 1--1.

\end{thebibliography}

\end{document}